\documentclass[oneside,a4paper]{book}

\usepackage{pdfpages}
\usepackage{caption}
\usepackage{amsthm}
\usepackage{xspace}
\usepackage{float}
\usepackage{ifthen}
\usepackage{amsbsy}
\usepackage{amssymb}
\usepackage{balance}
\usepackage{booktabs}
\usepackage{graphicx}
\usepackage{rotating}
\usepackage{multirow}
\usepackage{needspace}
\usepackage{microtype}
\usepackage{bold-extra}
\usepackage{geometry}
\usepackage{varioref}
\usepackage{xcolor}
\usepackage{textcomp}
\usepackage{listings}
\usepackage[normalem]{ulem} 
\usepackage{ucs}

\usepackage{times}
\usepackage{url}
\usepackage{alltt}
\usepackage{amsmath}
\usepackage{xfrac}
\usepackage{subfigure}
\usepackage{appendix}
\usepackage{stmaryrd}   
\usepackage[utopia]{quotchap}

\usepackage{setspace}
\usepackage[numbers, sort&compress]{natbib}
\usepackage{mdwlist}        
\usepackage{chngpage}
\usepackage[normalem]{ulem} 

\newboolean{showedits}
\setboolean{showedits}{true} 
\ifthenelse{\boolean{showedits}}
{
       \newcommand{\del}[1]{\textcolor{red}{\sout{#1}}} 
}{
       \newcommand{\del}[1]{} 
       
}

\usepackage{array}

\usepackage{pgfplots}
\pgfplotsset{compat=1.16}
\usepackage{xcolor}
\usepackage[normalem]{ulem}

\newboolean{showcomments}
\setboolean{showcomments}{true}
\ifthenelse{\boolean{showcomments}}
    {\newcommand{\nb}[3]{
        {\colorbox{#2}{\bfseries\sffamily\scriptsize\textcolor{white}{#1}}}
        {\textcolor{#2}{\sf\small$\blacktriangleright$\textit{#3}$\blacktriangleleft$}}}

	 \newcommand{\chk}[1]{\textcolor{ForestGreen}{#1}} 
	}
    {\newcommand{\nb}[3]{}
     
	\newcommand{\chk}[1]{} 
	}

\usepackage{xcolor}
\usepackage{textcomp}
\usepackage{listings}
\definecolor{source}{gray}{0.9}
\newcommand{\tabref}[1]{\hyperref[{tab:#1}]{Table~\ref*{tab:#1}}}
\newcommand{\figref}[1]{\hyperref[{fig:#1}]{Figure~\ref*{fig:#1}}}
\newcommand{\secref}[1]{\hyperref[{sec:#1}]{Section~\ref*{sec:#1}}}
\newcommand{\lstref}[1]{\hyperref[{lst:#1}]{Listing~\ref*{lst:#1}}}
\newcommand{\charef}[1]{\hyperref[{cha:#1}]{Chapter~\ref*{cha:#1}}}

\newcommand{\newevenside}{
	\ifthenelse{\isodd{\thepage}}{\newpage}{
	\newpage
        \phantom{placeholder} 
	\thispagestyle{empty} 
	\newpage
	}
}

\DeclareGraphicsExtensions{.pdf,.png}
\graphicspath{{images/}}

\newcommand{\thesistitle}{Using Word Embeddings to Analyze Protests News}
\newcommand{\thesisauthor}{Maria Alejandra Cardoza Ceron}
\newcommand{\thesisleitermain}{Prof. Dr. Martin Christian Braschler}
\newcommand{\thesisleiter}{Prof. Dr. Jacques Savoy}

\newcommand{\thesissubtitle}{}
\newcommand{\thesisdate}{September 2021}

\usepackage[ colorlinks=true, urlcolor=black, linkcolor=black,
			citecolor=black, bookmarksnumbered=true, bookmarks=true,
			plainpages=false,
			pdftitle={\thesistitle}, pdfauthor={\thesisauthor},
			pdfsubject={\thesissubtitle}, pdfpagelabels]{hyperref}

\newcommand{\hrref}[2]{\hyperref}

\begin{document}

\begin{titlepage}  
  \begin{center}  
  
  \begin{figure}[t]  
  \vspace*{-2cm}        
  \center{\includegraphics[scale=0.2]{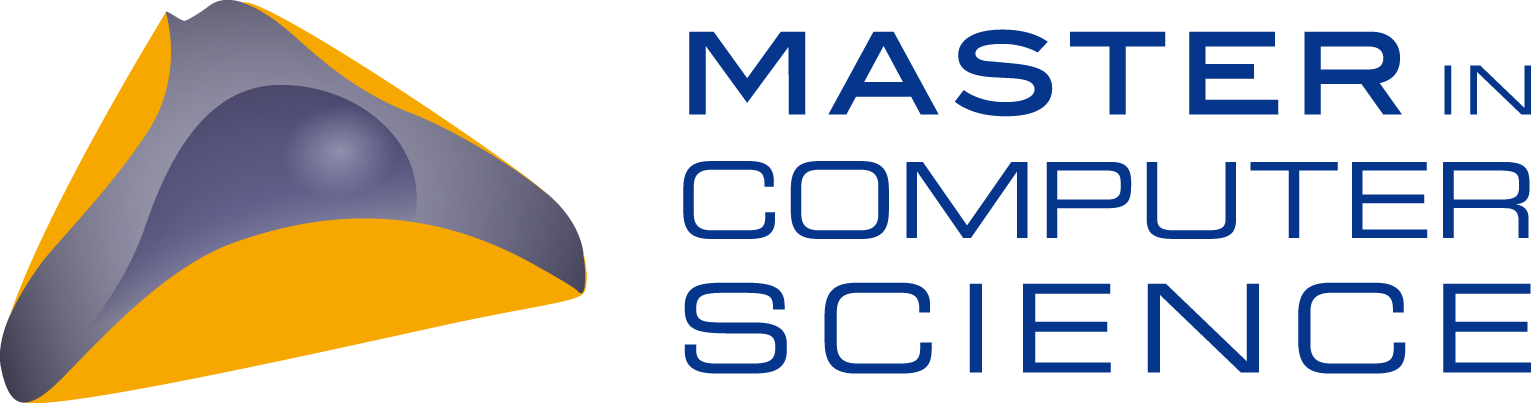}}
  \vspace{0.4in}     
  \end{figure}

    \thispagestyle{empty}
    
    {\bfseries\Huge \thesistitle \par
    \Large \vspace{0.1in} \thesissubtitle \par}

    \vspace{0.3in} 
    \LARGE{\textbf{Master Thesis} \\}
    \vspace{0.4in}

    {\Large \thesisauthor}
    
    \vspace{0.3in}
    {\Large University of Neuchâtel \par}
    \vfill
    {\Large \thesisdate \par}
    \vspace{0.3in}
    {\Large Supervisors \par}
    {\Large \thesisleitermain \par}
    {\Large \thesisleiter \par}
    {\Large Computational Linguistics \par}

  \vspace{0.9in}

  \begin{figure}[htp]
    \centering
    \includegraphics[scale=0.30]{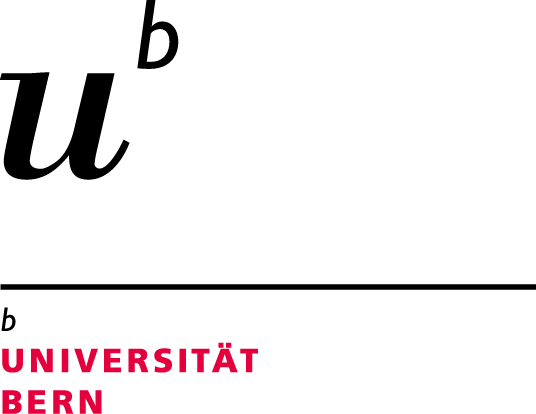}\hfill
    \includegraphics[scale=0.30]{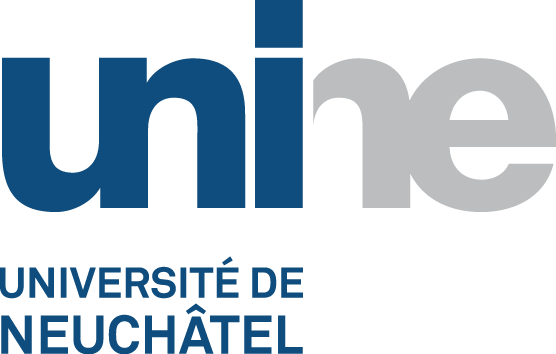}\hfill
    \includegraphics[scale=0.80]{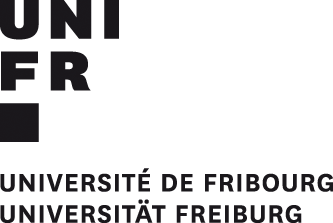}
  \end{figure}

  \end{center}

\end{titlepage}

\chapter*{\centering Abstract}
\begin{quotation}
\noindent 
The first two tasks of the CLEF 2019 ProtestNews events focused on distinguishing between protest and non-protest related news articles and sentences in a binary classification task. Among the submissions, two well performing models have been chosen in order to replace the existing word embeddings word2vec and FastTest with ELMo and DistilBERT. Unlike bag of words or earlier vector approaches, ELMo and DistilBERT represent words as a sequence of vectors by capturing the meaning based on contextual information in the text. Without changing the architecture of the original models other than the word embeddings, the implementation of DistilBERT improved the performance measured on the F1-Score of 0.66 compared to the FastText implementation. DistilBERT also outperformed ELMo in both tasks and models. Cleaning the datasets by removing stopwords and lemmatizing the words has been shown to make the models more generalizable across different contexts when training on a dataset with Indian news articles and evaluating the models on a dataset with news articles from China.

\end{quotation}
\clearpage

\tableofcontents

\listoffigures
\listoftables

\chapter{Introduction}
\label{chapter:introduction}
One of the most famous goals in computer science is to create a machine that passes the imitation game proposed by Alan Turing in 1950, today commonly known as the Turing-Test \cite{turing2009computing}. If a person can not reliably tell whether a conversation is happening with a machine or another person, the Turing-Test is said to be passed. While this is one of the earliest mentioning of artificial intelligence, it also opened up the research field of Natural Language Processing (NLP). NLP is concerned with the ability of a computer to analyze, process and understand natural language. Today, natural language processing is everywhere in the form of search engines, translations, chatbots, voice assistants, grammar checkers and many more. 

The difficulty for many NLP tasks is the understanding of natural language by an Articifial Intelligence (AI). When we read a text in a language we are fluent in, the process of understanding involves much more than just mapping a fixed meaning to a word. While it is certainly important to understand the meaning and part-of-speech of a single word, we also take into account the structure and meaning of the sentence the word appears in, the context in which the text was written, the intention of the author and many other information based on context and experience \cite{chowdhury2003natural}. It is no surprise that Natural Language Understanding is considered an AI-complete problem, as it can currently not be fully solved by an AI without human aid \cite{yampolskiy2013turing}.

Nevertheless, there have been many advances in NLP during the past years, with one of them being concerned about the representation of words in an unambiguous way, while still preserving the intended meaning. This representation is called word embedding, and usually represents the meaning of a word in form of a vector. Words with similar meanings get represented with vectors that are close in the vector space, giving the advantage of working with the meaning of words rather than the often times ambiguous words themselves \cite{jurafsky2014speech}.

One of the applications of word embeddings are in natural language classification tasks, where the meaning of words plays a central role. Classifying texts into different categories requires an understanding of what is written, rather than just recognizing individual words. Such a classification problem has been proposed in the CLEF ProtestNews Lab in 2019, with the goal of classifying texts stemming from news articles into the two categories protest-news related and protest-news unrelated\footnote{https://emw.ku.edu.tr/clef-protestnews-2019/}. The best results in the event have been achieved with the use of various word embedding methods \cite{crestani_overview_2019}.

Two novel word embedding methods called ELMo and DistilBERT use the contextual information from the text to map words to vectors. One of the main advantages of these two word embeddings is that the same word can be mapped to different vectors, depending on the meaning from the context \cite{peters2018deep}\cite{sanh2019distilbert}. This solves one of the difficulties when working with natural language, allowing ELMo and DistilBERT to perform very well in such classification tasks \cite{buyukoz_analyzing_nodate,buyukoz-etal-2020-analyzing}.

In the scope of this thesis, we pick two well performing models of the CLEF ProtestNews Lab and replace the existing word embeddings with ELMo and DistilBERT without changing the architecture of the models. The goal is to find out how well ELMo and DistilBERT perform when integrated into existing models. 

The next Chapter shows the related work to this thesis, including the two models used for our experiments. Chapter 3 explains the tasks and datasets used to evaluate the models. The preprocessing of the datasets and the architecture of the implemented models are described in Chapter 4. The evaluation and results of the experiments are found in Chapter 5, with the discussion interpreting the results following in Chapter 6. The last Chapter 7 concludes our work and shows how it can be continued and improved in the future.

\section*{Research Questions}
For our research questions we want to know if both ELMo and DistilBERT can be integrated into existing models as word embeddings to boost the performance for natural language classification tasks:
\begin{enumerate}
  \item Does the use of ELMo word embeddings increase the performance of an already established model using other word embeddings?
  \item Does the use of DistilBERT word embeddings increase the performance of an already established model using other word embeddings?
  \item Does ELMo or DistilBERT perform better measured on two well performing models in the CLEF 2019 Protest Event?
\end{enumerate}

\chapter{Related Literature}
Natural Language representation techniques used for Text Classification tasks (and many others) have been an interesting research topic during the last decades, with the goal of creating reliable and dynamic tools that can deal with textual information from different topics and sources.

Machine learning algorithms and neural networks in combination with word embedding techniques have proven to be useful in the development of different natural language tools, among them binary text classification tasks.

\section{Word embedding methods}

\subsection{Word2Vec} 
The word vector representation word2vec is an skip-gram model that uses neural networks for creating vector representations from the association of words in a large unsupervised corpus. Word2vec creates a unique continuous word representation for each unigram and bigram that have been found in certain document or sentence. The model maximizes the probability that two or more words appear together in one phrase as a representation of a bag of character n-grams \cite{mikolov2013distributed}.

\subsection{FastText} 
FastText is an n-gram word representation that handles subword information. The model is an extension of the continuous word vector representation model by taking into account the morphological structure of words such as suffixes and prefixes for modeling short and rare words much better. The approach can be used in supervised and unsupervised learning for different languages with different morphology forms achieving good results in text classification tasks \cite{bojanowski2017enriching}. 

\subsection{ELMo}
The ELMo embedding word representation from language models was introduced by M. Peters et al. in 2018 \cite{peters2018deep}. It takes both the syntactic and semantic information of a word into account and tries to solve the problem of representing words that share the same spelling but have many different possible meanings (homonyms). An example is the word \textit{letter}, which can both mean a written message to be sent to someone else or single character in the alphabet. This presents a big challenge in other embedding approaches, that only generate single contextualized representation of a word. ELMo on the other hand can represent the same word with different vectors depending on the context. The word vectors are the result of learned functions coming from the internal states of a deep bidrectional language model (biLM) that has been pretrained on a large corpus. While other word embedding methods only represent a token with a single layer in a fixed vocabulary, ELMo provides three layers of representations and can even represent words that did not appear in the training corpus solely relying on character input. The language model improves the state of the art of CoVe, word2vec and glove embedding models shown in six different natural language tasks, including text and sequence classification \cite{peters2018deep}.

\subsection{DistilBERT}
DistilBERT is a small Bidirectional Encoder Representation of BERT that retains 97\% of the BERT understanding abilities by using a knowledge distillation technique. DistilBERT is 40\% smaller and 60\% faster than BERT. The pretrained model helps to establish the context of words, trying to understand the meaning of unclear language in a text by using the surrounding text similar to ELMo. It also allows reasonable computational training for many different tasks by obtaining similar results as a normal Bidirectional Encoder Representations model. The implementation uses a tripe loss function, including a distillation loss and a cosine embedding loss. Its architecture remains similar to the BERT model except that DistilBERT does not make use of the token-type embeddings and the pooler, and the number of layers is reduced to 2 \cite{sanh2019distilbert}.

\section{Work based on the CLEF 2019 ProtestNews Event tasks}
Previous work has demonstrated that the use of pretrained contextualized language models can deal with the problem of complexity calculations that otherwise requires powerful hardware when the models need to be trained on a large dataset. It has also been established that pretrained models such as ELMo and DistilBERT perform very well in the development of NLP tools used for text and sentence classification task in the detection of ProtestNews locally and across different country domains.

\subsection{Attention Models for Text and Event Sentence Classification}
\begin{figure}[H]
    \centering
    \includegraphics[scale=0.6]{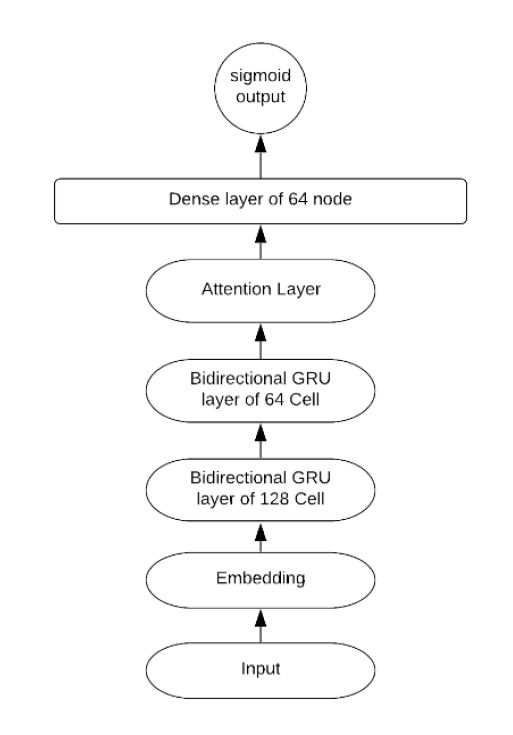}\hfill
     \caption{Deep learning models structure.}\label{SAF_Architecture}
  \end{figure}

The work of A. Safaya for the CLEF 2019 ProtestNews event will be the basis for the first one of our models \cite{safaya_event_nodate}. The focus of the model was the binary classification of sentences into protest news related sentences and any other news with the best overall F1-score 0.69 Figure \ref{Overview results CLEF 2019 ProtestsNews}, and the best results for task 2 of the event (Figure \ref{Overview results CLEF 2019 ProtestsNews}). The model uses Attention Models with Bidirectional Gated Recurrent Networks (GRU) and a word embedding layer based on word2vec. Figure \ref{SAF_Architecture} shows the architecture of the model. Word2vec, a word based representation embedding method, replaces each word with embedding vectors in the embedding layer. Due to the state-of-the-art performances of Recurrent Neural Networks (RNN), two bidirectional GRUs follow with 128 and 64 Cells. This is followed by an Attention Layer with context, which helps by putting more focus on words that are important to the context. Next is a dense layer with 64 nodes and ReLU activation function. The output layer using a sigmoid function has one node for the final binary classification. In our case, we will implement this model for task 1 and task 2 replacing the word2vec word embedding layer with ELMo and DistilBERT respectively, while keeping the rest of the model the same. This will allow us a direct comparison of these word embedding methods.

\subsection{Multitask Models for detecting ProtestNews in Texts}
\begin{figure}[H]
    \centering
    \includegraphics[scale=0.6]{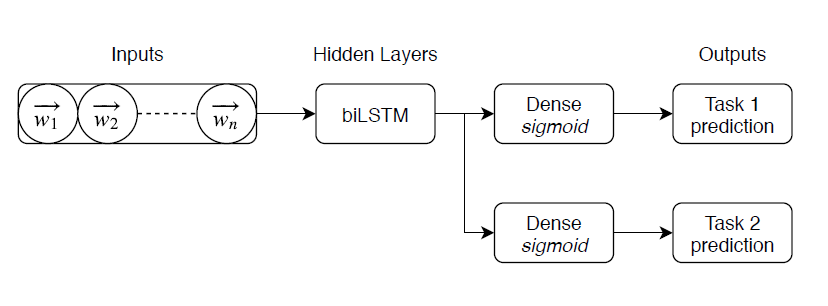}\hfill
     \caption{Multitask Model architecture for task 1 and 2.}\label{BEN_Architecture}
  \end{figure}

B. Radford proposed a Multitask Neural Network model that outputs simultaneous predictions for both the binary text classification of task 1 and the event sentence classification for task 2 \cite{radford_multitask_nodate}. The model achieved the best results for task 1 and the third best results overall with an average F1-Score of 0.65 for task 1 and task 2 (Figure \ref{Overview results CLEF 2019 ProtestsNews}). It will be used in this work as the second base model for the implementation of ELMo and DistilBERT. The architecture proposed for both tasks uses a Recurrent Neural Network as seen in Figure \ref{BEN_Architecture}. On the left, the sequence input is shown for the word vectors, the first layer of the model contains 10 biLSTM (Bidirectional long short term memory) Recurrent Neural Networks cells without an activation function. The activation values of the cells at the final sequence token of this layer are fed to two fully-connected dense layers. One of these dense layers is trained on the dataset from task 1 (documents), the other dense layer is trained on the dataset from task 2 (sentences). The sigmoid function of these layers maps the output between 0 and 1 to protest and non-protest. The fitting is done with RMSProp and log loss is used for the loss function. The word vectors for the embeddings were pretrained using FastText, due to the size of the data and limited computing resources.

\subsection{Further work}  
\begin{figure}[H]
    \centering
    \includegraphics[scale=0.8]{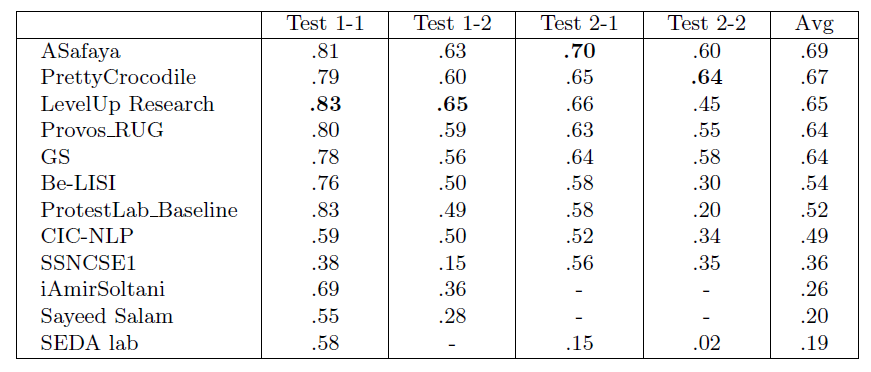}\hfill
     \caption{Overview of the ranking results of all the participant teams on the ProtestNews 2019 Lab for task 1 and task 2  \cite{crestani_overview_2019}.}\label{Overview results CLEF 2019 ProtestsNews}
\end{figure}

ELMo word representations for news protest classification has been used by E. Maslennikova, under the team name of PrettyCrocodile. The model got the second best average result of 0.67 in the two tasks of the CLEF 2019 ProtestNews event seen in Figure \ref{Overview results CLEF 2019 ProtestsNews}, which included document text classification and event sentence detection tasks\cite{maslennikova_elmo_nodate}. The best model for task 1 was implemented using a pretrained ELMo model in combination with a fully connected neural network containing 4 Dense layers. The best model for task 2 used ELMo and 6 Convolutional and MaxPooling layers, combined with GlobalAveragePooling, Dropout and Dense layers. 

Pretrained word embeddings such as FastText in combination with a basic logistic regression classifier were implemented for identifying protest events in news articles by A. Basile et al. \cite{basile_protesta_nodate}. The architecture proposed for task 1 and task 2 is a stacked ensemble model of logistic regression classifiers, that uses a pretrained FastText embedding with 300 dimensions utilizing sub-word information. The model was ranking fourth on the average results of the CLEF 2019 Lab with an F1-Score of 0.64 Figure \ref{Overview results CLEF 2019 ProtestsNews}.

Different pretrained word embedding approaches such as word2vec, GloVe and FastText together with support vector machines and two different linear classifiers were proposed by A. Ollagnier et al. in the development of the protestnews lab \cite{ali_linguistic_nodate}. The model uses the three word-embedding techniques mentioned before in combination with linear classifiers such as support vector machines and a decision tree boosted with XGBoost. It also utilizes a singular value decomposition technique for the dimensional reduction process. The SEDA lab team got the last position on the Lab results for task 1 and task 2 with an average f1-score of 0.19, seen in Figure \ref{Overview results CLEF 2019 ProtestsNews}.

The contextualized string embeddings Glove and FastText were stacked for developing the binary text and sentence detection tasks by G. Skitalinskaya et al. \cite{skitalinskaya_clef_nodate}. The approach uses the contextualized word embeddings together with a pretrained Flair language model for generating the word vectors. This representation is used for the input sequence of the LSTM-based model. For classifying the documents a linear layer was implemented. Two experiments were run for task 2 with different batch sizes of 16 and 8 separately. The experiment number two showed that small batch sizes are much more noisy than larger ones. The model reached the third position for task 2 and the fifth place overall with an average F1-Score of 0.64 (Figure \ref{Overview results CLEF 2019 ProtestsNews}).

Supervised machine learning classifiers based on logistic regression algorithms combined with different linguistic parameters for the detection of protest news were implemented by A. Chedi Bechikh, a participant of the Be-LISI team\cite{ali_linguistic_nodate}. Document expansion with word embedding similarity were also implemented to deal with the term mismatch problem presented in task 2. The Be-LISI team ran 5 different experiments for the above task and combined different natural language techniques such as lemmatization, removal of stop words and named entities, and annotating all compound nouns. The model reached an average result of 0.54 in the protestnews lab (Figure \ref{Overview results CLEF 2019 ProtestsNews}).

Attention mechanisms with Long short-term memory (LSTM) models and Recurrent Neural Networks (RNN) were the focus of the SSNCSE1 team on the Lab. The work developed by D. Thenmozhi et al. proposed 4 different deep learning models with different attention mechanisms for task 1 and task 2 \cite{thenmozhi_extracting_nodate}. The approaches were implemented with a multi-layer recurrent neural network with two bidirectional LSTM layers with Bahdanau and Normed-Bahdanau attention and Luong and Scale Luong attention, one embedding layer, encoding and decoding layers, projection and loss layers. The best model got a F1 macro average score of 0.36 (Figure \ref{Overview results CLEF 2019 ProtestsNews}).

Further work that has been done one year later after the CLEF 2019 ProtestNews Lab, has demonstrated that the use of ELMo and DiltilBERT for test classification generalizes in average much better than the majority of the models proposed by the participants in the Lab. In the work of B. Buyukoz's, the ELMo model was used with two Feed Forward Neural Networks (FFNN) and DistilBERT in combination with two Bidirectional Long Short Term Memory networks (BiLSTM) to classify protestnews for task 1 \cite{buyukoz_analyzing_nodate}. The 2 classifiers were trained for 10 epochs with the Adam optimizer, using the linear activation function RELU, a dropout layer to prevent overfitting, and a softmax function that receives the vector predictions containing the logits function. The DistilBERT approach for text classification task got the best results on average with an F1-score of 0.79 for task 1, which is highest compared to the Ranking results of the ProtestNews Lab. The ELMo model got a macro average F1-score of 0.78 for task 1 being the second highest result on the CLEF lab ranking.

Furthermore, Sequential Transfer Learning (STL) techniques using various pretrained language models were implemented by A. Ollagnier et al. \cite{ollagnier_sequential_2020}. The approach uses seven different embedding architectures, among them BERT, DistilBERT and GPT-2 for the model development of task 1 and task 2. The model takes an input sequence of words than are put into the embedding layer, after that an 1D convectional neural network classifier is introduced with a filter of 32, kernel 3 and a ReLU linear activation function. An LSTM layer with 100 units is added, followed by two dense layers with 64 and 1 units, one with a ReLu and the other one with a sigmoid activation function that maps the prediction outputs for the binary classification task. The Adam optimizer with a learning error rate of 0.00005 and a batch size of 16 has been used. The results of the 9 different experiments show that the OpenAI GPT-2 embedding model outperforms the other ones on average with an 0.63 f1-score. Compared with the ranking results of the CLEF the model this model would be ranked in the sixth position.

\chapter{Tasks and Dataset}
The work in this thesis is based on the CLEF 2019 ProtestNews event\footnote{https://emw.ku.edu.tr/clef-protestnews-2019/} and will contain comparisons to previous results thereof. In order to get a fair comparison, we adopt the same tasks and datasets that have been used in the event. The main idea is to extract information about potential protests from English news articles of different countries, namely China and India. The goal of the event was to enhance different machine learning approaches with word embedding techniques in order to see if this results in a good performance of the models. Both task 1 and 2 are binary classification problems whether the sentence or text contain information about protests \cite{crestani_overview_2019}. This chapter will provide an overview of the two tasks we are interested in and some analysis about the datasets.

\section{Task 1: News Text Classification (Protest vs. Non-Protest)}

The goal of task 1 is to classify English news articles from India and China into protest event related news and non-protest news. In addition to the machine learning aspect of this task, at least one word embedding method should be used. The participants were free to apply any other natural language processing techniques such as preprocessing the news article texts for example by removing special characters or stemming the words. 

The organizers of the CLEF ProtestNews event 2019 provided multiple datasets for this task:
\begin{enumerate}
  \item A large, labelled \textbf{training dataset} consisting of English news articles from India
  \item A smaller, labelled \textbf{development dataset} for trying out models or verifying the performance a trained model also consisting of English news articles from India
  \item An unlabelled \textbf{testing dataset} with English texts from Indian news articles\footnote{The labels for the test datasets are hidden to the participants, but predictions can be uploaded to get performance scores}
  \item An unlabelled \textbf{testing dataset} with English texts from Chinese news articles\footnote{See footnote 2}
\end{enumerate}

We can see that for training the models we only have access to labelled English news articles from India, but one of the testing datasets is from a different country, China. The reason behind this is to find out if models can perform well across texts from different sources, i.e. news articles from different countries. 

\section{Task 2: Event Sentence Detection}

The goal of task 2 is to classify single sentences from news articles into whether they contain information about a protest event or not. This task is similar to task 1, with the only difference being that we only have one sentence instead of the whole news article. Due to being limited to a single sentence, this task is generally more difficult than task 1.

The datasets provided for this task are the following:
\begin{enumerate}
  \item A large, labelled \textbf{training dataset} consisting of English sentences extracted from news articles from India
  \item A smaller, labelled \textbf{development dataset} for trying out models or verifying the performance a trained model also consisting of English sentences extracted from news articles from India
  \item An unlabelled \textbf{testing dataset} with English sentences from Indian news articles
  \item An unlabelled \textbf{testing dataset} with English sentences from Chinese news articles
\end{enumerate}

Again, we only have labelled datasets from India, but one of the testing datasets is from Chinese news articles.

\section{Dataset}

The ProtestNews datasets were aquired from online English news articles from India and China. The binary labelling into protest and non-protest news articles or sentences respectively has been done by hand for both the training and testing datasets \cite{azzopardi_task_2019}, which is created using random sampling \cite{yoruk+2021} and cross-context \cite{hurriyetoglu+2021} setting. 

We will refer to the different datasets as follows:
\begin{itemize}
  \item train\_india: The Indian training dataset containing labelled texts (task1) or labelled sentences (task2)
  \item dev\_india: The smaller Indian development dataset, containing labelled texts (task1) or labelled sentences (task2)
  \item test\_india: The Indian testing dataset, containing unlabelled texts (task1) or unlabelled sentences (task2)
  \item test\_china: The China testing dataset as the cross-context, containing unlabelled texts (task1) or unlabelled sentences (task2)
\end{itemize}

Tables \ref{table:task1} and \ref{table:task2} show an overview of all the datasets. We can see that the datasets are not balanced, as there are more protest news articles and sentences compared to non-protest entries. For task 1 we get a protest-to-non-protest ratio in the train\_india dataset of 0.22 from equation \ref{equation:eq1}. 

\begin{equation} 
\label{equation:eq1}
\begin{split}
Ratio & = \frac{Protest\_texts}{Total\_texts} 
\end{split}
\end{equation}

For task 1 we can also see that the ratio for the india\_test dataset stays the same 0.22. However, we have a much more imbalanced test\_china dataset with a ratio of 0.05, meaning the vast majority of news articles are not about protests. We got this information about the test datasets from INSERTREFERENCEHERE, and it will not be used in the building of the models as the participants in the original CLEF 2019 event did not have access to this information.

For task 2 we notice a similar imbalance, with a protest-to-non-protest ratio of 0.17 for the train\_india dataset and 0.21 for the dev\_india dataset. We can assume that this ratio stays roughly the same for the test\_india dataset, but we have no information on the imbalance of the china\_test dataset.

\begin{table}[H]
\begin{center}
\centering
\caption{Data samples distribution for task 1.}
\label{table:task1}
 \begin{tabular}{ |c|c|c|c|c| } 
 \hline
 \multicolumn{5} { |c| } {Task 1: Binary Text Classification} \\
  \hline
 \textbf{Document Datasets} & \textbf{Size} & \textbf{Protest} & \textbf{Non-Protest} & \textbf{Protest\_Ratio} \\ 
 \hline
 \textbf{Train\_India} & 3430 & 769 & 2661 & 0.22 \\ 
 \hline
 \textbf{Dev\_India} & 457 & 102 & 355 & 0.22 \\ 
 \hline
 \textbf{Test\_India} & 687 & 15$1^1$ & 536 & 0.22$^1$ \\ 
 \hline
 \textbf{Test\_China} & 1801 & 9$0^1$ & 1711 & 0.0$5^1$ \\ 
 \hline
\end{tabular}

$^1$We got the ratios for the test datasets from \cite{buyukoz_analyzing_nodate}.
\end{center}
\end{table}

\begin{table}[H]
\centering
\caption{Data samples distribution for task 2.}
\label{table:task2}
\begin{center}
 \begin{tabular}{ |c|c|c|c|c| } 
 \hline
 \multicolumn{5} { |c| } {Task 2: Event Sentence Detection} \\
  \hline
 \textbf{Sentence Datasets} & \textbf{Size} & \textbf{Protest} & \textbf{Non-Protest} & \textbf{Protest\_Ratio} \\ 
 \hline
 \textbf{Train\_India} & 5885 & 988 & 4897 & 0.17 \\ 
 \hline
 \textbf{Dev\_India} & 663 & 138 & 525 & 0.21 \\ 
 \hline
 \textbf{Test\_India} & 1107 & * & * & * \\ 
 \hline
 \textbf{Test\_China} & 1235 & * & * & *\\ 
 \hline
\end{tabular}

*This information was kept hidden by the organizers of the ProtestNews lab.
\end{center}
\end{table}

\subsection{Detailed dataset analysis Task 1} 

Table \ref{table:exampletask1} shows two excerpts of the train\_india dataset with one protest and one non-protest news article. We can clearly see the opposing sentiment of these two examples. The first, protest news related article has a very negative sentiment containing words such as 'shock', 'grief', 'gruesome' or 'condemning'. Additionally, it contains references to politics such as the Chief Minister or the political party involved in the protest. The second, non-protest news article has a very positive sentiment about a celebration with no references to politics. This is where word embedding comes into play, as it facilitates the differentiation between such articles even without solely relying on single words, but also taking the context and general sentiment into account to capture the meaning \cite{peters2018deep} \cite{sanh2019distilbert}. This is especially important when having test datasets from different sources, in this case different countries, as the formulation and used words may not be very similar \cite{ettinger2017towards}.

\begin{table}[H]
\begin{center}
\centering
\caption{Example of a news article for task 1.}
\label{table:exampletask1}
 \begin{tabular}{|p{10cm}|c|}
 \hline
 \textbf{Text} & \textbf{Label} \\ 
 \hline
 [...] haryana chief minister om prakash chautala has expressed shock and grief over the gruesome killing of ms phoolan devi, mp. chautala said that phoolan devi, who belonged to samajwadi party, had carved a niche for herself in country's political circles in a short period of her political career. condemning the killing, he [...] & Protest \\ 
 \hline
 [...] The authentic feel and aura of Santiniketan came alive in Helen Jerwood Hall of St. Thomas School here over the weekend as it celebrated the 150th birth anniversary of Rabindranath Tagore. Vignettes from Tagore's outstanding plays [...] & Non-Protest \\ 
 \hline
\end{tabular}
\end{center}
\end{table}

Table \ref{table:task1_detailed} shows some interesting differences between the texts from the Indian news articles and the Chinese news articles.

The average text length shows how many tokens (words, numbers, special characters) on average a text of the different datasets has. The texts of the test\_china dataset are with 331 tokens on average slightly larger than the Indian datasets with a length between 307 and 312. The number in the brackets shows the standard deviation normalized by N-1. We can see that the length of the texts from test\_china vary much more than the Indian texts, with a standard deviation of 251 compared to 172 (test\_india), 183 (train\_india) and 190 (dev\_india). 

Following up on that, we calculated the number of 'short texts' for each datasets, which are texts with fewer than 100 tokens. Unsurprisingly, the china\_test dataset has the highest amount of short texts with 11\%. In comparison, the Indian dataset only contains between 7-8\% texts with fewer than 100 tokens. 

The sentences of the Chinese datasets are also longer with an average of 37 tokens, compared to 27 of test\_india and 28 of train\_india and dev\_india. 

As a last measurement, we calculated the Lexical Density for all the datasets. The Lexical Density shows us how dense the information in a text is by dividing the number of lexical words (nouns, verbs, adjectives and adverbs) by the total number of tokens. The Lexical Density is very similar across all datasets with about 0.57 each. Together with the average text length, standard deviation and number of short texts, this shows us that even though the texts of the test\_china dataset contain more information on average, there is also a higher number of texts that contain less information compared to the Indian datasets. Together with the possibly different words and formulations, this could pose a challenge to a trained model on the india\_train dataset when doing predictions on the test\_china dataset.

\begin{table}[H]
\begin{center}
\centering
\caption{Various measurements on the datasets of task 1.}
\label{table:task1_detailed}
 \begin{tabular}{ |c|c|c|c|c| } 
 \hline
 \textbf{Datasets} & \textbf{Avg. text length (Std. Dev.)} & \textbf{Short texts} & \textbf{Mean sentence length} & \textbf{Lexical Density} \\ 
 \hline
 \textbf{Train\_India} & 307 (+/-183) & 0.07 & 28 & 0.5710 \\ 
 \hline
 \textbf{Dev\_India} & 316 (+/-190) & 0.08 & 28 & 0.5696 \\ 
 \hline
 \textbf{Test\_India} & 312 (+/-172) & 0.07 & 27 & 0.5705 \\ 
 \hline
 \textbf{Test\_China} & 331 (+/-251) & 0.11 & 37 & 0.5714 \\ 
 \hline
\end{tabular}
\end{center}
\end{table}

\subsection{Detailed dataset analysis Task 2}

Table \ref{table:exampletask2} shows two example sentences, one protest and one non-protest, from the train\_india dataset of task 2. We can immediately see that we have much less information to work with compared to task 1. The first sentence, labelled as protest, contains direct information of a past protest and its outcome. The second sentence, labelled as non-protest, contains information about residents complaining about various subjects. However, this sentence does not directly contain any information about an upcoming or past protest. Nevertheless, the models need to be able to differentiate between the few features available per sentence. Sometimes, a single word can make the difference whether it is a protest or not. It is also important to mention that non-protest sentences may come from a protest related newspaper article.

\begin{table}[H]
\begin{center}
\centering
\caption{Example of news article sentences for task 2.}
\label{table:exampletask2}
 \begin{tabular}{|p{10cm}|c|}
 \hline
 \textbf{Sentence} & \textbf{Label} \\ 
 \hline
 The Maoists hit back with a vengeance within 48 hours, butchering 76 security personnel  & Protest \\ 
 \hline
 The residents complained about bad roads, water problems, shortage of power and others  & Non-Protest \\ 
 \hline
\end{tabular}
\end{center}
\end{table}

As with task 1, Table \ref{table:task2_detailed} further information about the different datasets. As the lexical density is not so interesting for single sentences, we instead analyze the number of stopwords and special characters that are not so useful for the classification. If these make up a large part of the sentence, the model does not have a lot of information to work with.

\begin{table}[H]
\begin{center}
\centering
\caption{Various measurements on the datasets of task 2.}
\label{table:task2_detailed}
 \begin{tabular}{ |c|c|c|c|c| } 
 \hline
 \textbf{Datasets} & \textbf{Avg. text length (Std. Dev.)} & \textbf{Short texts} & \textbf{Avg. Stopwords} & \textbf{Avg. Special Chars} \\
 \hline
 \textbf{Train\_India} & 24.1 (+/-13.0) & 0.08 & 9.97 & 2.61 \\ 
 \hline
 \textbf{Dev\_India} &  25.3 (+/-13.6) & 0.08 & 10.11 & 2.80 \\ 
 \hline
 \textbf{Test\_India} & 24.2 (+/-11.9) & 0.08 & 10.01 & 2.68 \\ 
 \hline
 \textbf{Test\_China} &  23.8 (+/-13.8) & 0.12 & 9.81 & 2.38 \\ 
 \hline
\end{tabular}
\end{center}
\end{table}

In the first column we can see the average text length each dataset has. Unlike the datasets in task 1, the test\_china dataset has the shortest sentence length on average with 23.8 tokens, but again also the highest standard deviation with 13.8 tokens (again normalized by N-1.). This means we have on average less information to work with in the test\_china dataset compared to the training and testing sets from India.

The column \textit{Short texts} shows the relative amount of sentences with less than 10 tokens. We can see that 12\% of the sentences in the test\_china dataset have less than 10 tokens, compared to 8\% for the datasets from India. This confirms further that we have less information to make the predictions for the test\_china dataset.

The two last columns, \textit{Avg. Stopwords} and \textit{Avg. Special Chars} show the average numbers of stopwords and special characters per sentence in the datasets. Stopwords are the most common words in a given language which do not contain much information and thus are often times filtered out when cleaning a dataset in a natural language classification task. 

We can see that the test\_china dataset contains the lowest average number of stopwords and special characters compared to the indian datasets. This means that even though the sentences are smaller in the test\_china dataset, we also have a bit more information to work with. For example if we remove all the stopwords and special characters during the data cleaning process, on average more tokens will remain in the sentences of the test\_china dataset.

\chapter{Methodology}
This chapter gives an overview of the methods used for the implementation of the different models of task 1 and 2. The first section, \ref{section:data_preprocessing} Data Preprocessing, shows the methods we have used to preprocess the different datasets. \ref{section:architectures} Architectures contains information about the ELMo and DistilBERT embedding layers and the structure of the different models. The last two sections, \ref{section:implementation_of_the_models} Implementation of the Models and \ref{section:training_the_models} Training the Models, give insights of the concrete implementation in python and the infrastructure used to process the models.

\section{Data Preprocessing}
\label{section:data_preprocessing}

As with all classficiation tasks in machine learning, it is important to understand the datasets used to train the models and make the classifications. A model can only be as good as the dataset used to train it. In natural language processing tasks especially there are many techniques that can be used to remove noise from the texts which can lead to a better performance for the predictions. However, there is no "one-fits-all" method for preprocessing the data. Some models may work very well without any or only slight alterations of the texts, while others may profit from a combination of different methods. This is why we have decided to run all of our models without any data preprocessing, with some light data cleaning and with a more extensive preprocessing of the data (the latter two will be discussed in this section). This allows us to experimentally find out what works best.

\subsection{Stopwords and special characters}
When classifying a text into different categories, not all words are equally useful. One might think that the most frequently appearing words are the most useful for gathering information. When analyzing these words one finds quickly that these are predominately function words used to give a sentence structure and only convey little information. Lexical words such as nouns, verbs, adjectives and adverbs on the other hand contain the most information \cite{freeborn1987lexical}.

A machine learning algorithm does usually not have this information and uses all tokens equally as predictors for a certain class. This means that it can happen that a function word is wrongfully used as a means of classification, even when it does not have an apparent link to the class. This can happen at random, for example when the texts used to train an algorithm just happen to have certain function words appear more often in a certain class. When testing said algorithm on a different dataset, this information may again be used for classification and lower the performance.

The easiest and probably most frequently used method to tackle this problem is to remove the most frequent appearing words in the English language from all datasets. Such words are called stopwords and there are many lists containing various amounts of stopwords that can be used to preprocess a dataset. We have used the stopwords list from the NLTK corpus\footnote{Natural Language Toolkit: https://www.nltk.org/, the list of the stopwords used in this work can be found here: https://www.nltk.org/nltk\_data/}, containing words such as \textit{I}, \textit{a}, \textit{be} or \textit{of}.

Even though these stopwords contain very little information by themselves, in some cases they can still be helpful for classifications. For this reason we have trained and tested all of our models with an untouched dataset where these stopwords are not filtered out, in addition to the two different cleaned datasets.

Similarly, we have found that also removing special characters instead of only stopwords boosts the performance of our models in most cases.

\subsection{Stemming / Lemmatization}
Another frequently used technique when cleaning texts is stemming or lemmatization. Stemming or word normalization reduces words to their root form or stem. This is useful, because the English language has many inflected and derived words that still have a similar meaning. Especially in cases with a limited amount of words as we have in Task 2, inflected words will be grouped together by their stem as a single predictor. 

However, the stem does not need to be a valid word, for example the words \textit{library}, \textit{librarian} and \textit{libraries} can get reduced to the common stem \textit{librari}. This is a problem for us, since we use word embeddings. Both ELMo and DistilBERT need to have valid words as inputs to capture the meaning of the text. Fortunately, there is a stemming technique called lemmatization which always returns valid words. Lemmatization reduces words to their dictionary form and can be used in combination with word embeddings, as the meaning of the word is not lost.

\subsection{Light cleaning}
Our first text cleaning method will be referred to as \textit{Light cleaning}. As the name implies, we only do a minimal cleaning compared to the original text:
\begin{itemize}
  \item Remove all stopwords
  \item Remove all special characters
  \item Transform all words to lowercase
\end{itemize}

\subsection{Cleaning}
Our second text cleaning method will be simply referred to as \textit{Cleaning}. Compared to the light cleaning, we additionally lemmatize all words and remove all proper nouns such as the names of cities or people. As we will see in the next chapter, this can reduce the performance of our models as some information is lost when cleaning the texts. For the lemmatization this can happen when the inflected words that get reduced to their dictionary form have a slightly different meaning in context. Removing proper nouns can also hurt the predictions, for example when a certain name is often mentioned in conjunction with protests and is used as a predictor. On the other hand it can also make the model more generalizable and in the case of proper nouns it will no longer be as dependant on specific locations or people.

Cleaning the dataset is composed of:
\begin{itemize}
  \item Remove all stopwords
  \item Remove all special characters
  \item Transform all words to lowercase
  \item Lemmatize all words
  \item Remove all proper nouns
\end{itemize}

\subsection{Dataset specific cleaning}
During the evaluation of our models we also took a closer look at the misclassifications to find out what might be the reason that some texts have been assigned to the wrong class. We have found that some texts in our Indian datasets for task 1 did not only contain the newspaper article that should be classified, but also the title of some related articles. This is the result of extracting the newspaper articles from websites where these related articles were wrongfully extracted in addition to the actual title and text of the article that should be used for the classification.

We decided to delete this unwanted information and found an overall increase in performance of the models. As this is more of a problem on how the datasets were obtained, rather than a cleaning step of the intended texts, we treated these datasets without any information of related articles as our new datasets. So all the models we ran used these new datasets, even the ones denoted with \textit{NotClean} in chapter \ref{chapter:evaluation_and_results} Evaluation and Results.

\subsection{SMOTE: Synthetic Minority Over-sampling Technique}
A problem we faced when replacing the existing word embedding methods with ELMo and DistilBERT was the prediction of only one class. The reason for this is the imbalanced dataset used to train the model, with around 80\% of all newspaper articles (task 1) and sentences (task 2) belonging to the non-protest news class. When trying to fit the model to the training dataset, the loss of the model often times did not converge to 0. As a result, the model always predicted the class non-protest.

There are a few ways to fix that problem, but since we do not want to change the model itself we are limited to changing the training dataset. We decided to resample the training dataset to get a more balanced representation of the two classes. There are two resampling techniques: oversampling and undersampling. Undersampling is the reduction of the majority class, non-protest in our case, and is generally used for big datasets. As our trainig datasets consist of only 769 minority class samples for task 1 and 988 minority class samples for task 2, we decided to use oversampling instead. As the name implies, oversampling is about generating more samples for the minority class. 

We have found that SMOTE, the synthetic minority oversampling technique, worked best for our training datasets. SMOTE chooses a random sample from the minority class and then finds the k-nearest neighbors (5 in our case) for that sample. Next, a random sample from the 5 nearest neighbors is chosen and a synthetic sample from the original and the chosen neighbor is generated \cite{chawla2002smote}. 

SMOTE is applied after the word embeddings as they utilize context information which will not necessarily be available for the synthetic samples. We apply SMOTE for the DistilBERT word embeddings for both model 1 and model 2. 

\section{Architectures}
\label{section:architectures}
\subsection{ELMo and DistilBERT Architecture for Model 1}
\label{subsection:model1:architecture}

\begin{figure}[hbt!]
    \centering
    \includegraphics[scale=0.5]{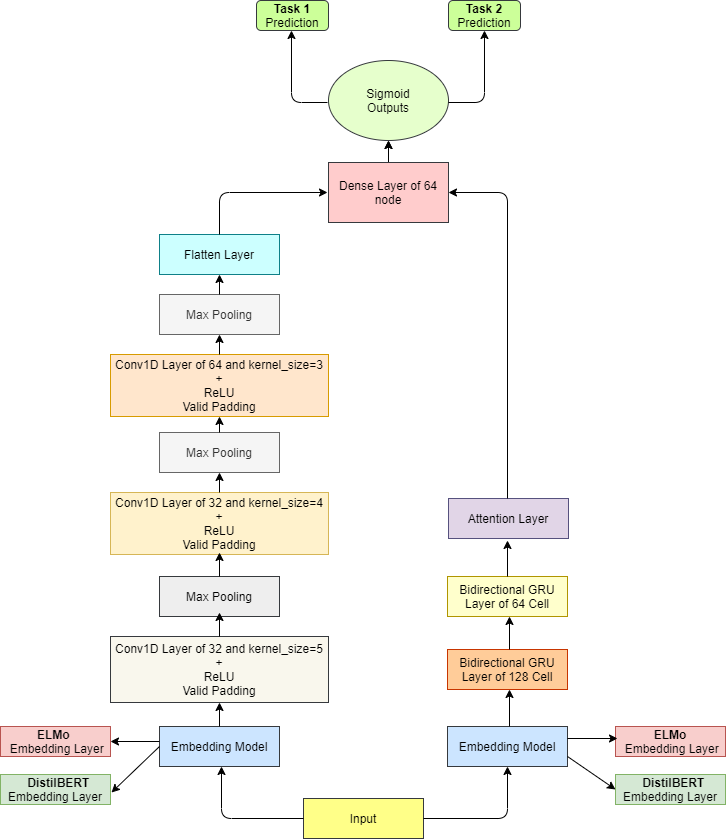}\hfill
     \caption{Architecture Model 1 for Task 1 and Task 2.}\label{ELMO-DistilBERT_Architecture_SAF}
\end{figure}

An overview of the architecture for the first model can be seen in Figure \ref{ELMO-DistilBERT_Architecture_SAF}. The input layer takes 800 tokens (padded or split where necessary) for task 1 and 100 tokens for task 2. The embedding layers consisting of the pre-trained ELMo or DistilBERT contextualized models output 256 embedded vectors derived from the input for task 1 and 32 embedded vectors for task 2. 

The next layer for task 1 is a one-dimensional convolution layer of 32 output filters (dimensionality of the output space) with a kernel size of 5 (length of the convolution window). The next layer, a one-dimensional Max Pooling layer downsamples the input with a window size of 3. The next layers are two 1D convolution layers with 32 filters/kernel size of 4 and 64 filters/kernel size of 3 with each having a Max Pooling layer with window size 3 as the next layer. The tensors then get reshaped in the Flatten layer before being fed to a Dense layer of 64 nodes. Finally, the sigmoid activation function converts the inputs to values between 0 and 1 corresponding to non-protest and protest articles \cite{safaya_event_nodate}.

Instead of convolution layers, the model for task 2 uses Bidirectional Gates Recurrent Units (GRU) instead. The output of the embedding layer is used as input by two sequential Bidirectional GRU layers of 128 and 64 cells. Next comes an Attention with Context layer, the output of which is again fed to a Dense layer of 64 nodes before being converted to an output between 0 and 1 with the sigmoid activation function corresponding to the two classes non-protest and protest \cite{safaya_event_nodate}.

\subsection{ELMo and DistilBERT Architecture for Model 2}
\label{subsection:architecture_model2}

\begin{figure}[hbt!]
    \centering
    \includegraphics[scale=0.5]{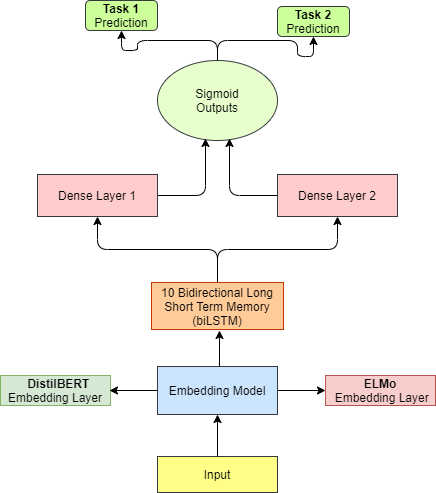}\hfill
     \caption{Architecture Model 2 for task 1 and 2.}\label{ELMO-DistilBERT_Architecture_BEN_architecture}
\end{figure}

The neural networks architecture Figure \ref{ELMO-DistilBERT_Architecture_BEN_architecture} implemented for Model 1 in this work, is a multitask learning process where the model is trained on multiple tasks at the same time. The first layer of the model is again the embedding layer consisting of the pre-trained ELMo or DistilBERT contextualized models. The embedding layer again takes 256 vectors derived from the 800 input tokens for task 1 and 32 vectors derived from the 100 input tokens for task 2. The second layer consists of a Bidirectional Long Short Term Memory (biLSTM) of 10 cells without an activation function. The outputs of the biLSTM layer are fully-connected with the next two dense layers where they compute the weighted sum of the the 10 obtained activation values. The first dense layer is trained only with protestnews articles from task 1 and the second one is trained on protestnews sequences of articles from task 2. The sigmoid activation function limits the outputs of the two dense layers to a range between 0 and 1 that correspond to non-protest and protest respectively \cite{radford_multitask_nodate}.

\subsection{ELMo Model}
For our experiments with ELMo we used the ELMo Hub module (v2) hosted on the TensorFlow Hub\footnote{https://tfhub.dev/google/elmo/2}. This model has been pretrained on the 1 Billion Word Benchmark Dataset. The use of pretrained word representation models has the advantage of not relying on corpus size or computational resources, while still being able to tweak the model \cite{mikolov2017advances}. 

For finetuning the ELMo model exposes 4 scalar weights that can be trained, while all other parameters remain fixed \cite{peters2018deep}. 

The ELMo model is integrated as an embedding layer into our models and receives the preprocessed texts as an input. The output is the weighted sum of the 3 ELMo layers word\_emb (character-based word representations), lstm\_outputs1 (the first LSTM hidden state) and lstm\_outputs2 (the second LSTM hidden state).

\subsection{DistilBERT Model}
For DistilBERT we used the pretrained uncased base model\footnote{https://huggingface.co/distilbert-base-uncased}. As this model is a distilled version of the BERT base model, it has been trained on the same corpus of 11,038 unpublished books and the English Wikipedia. DistilBERT has 40\% less parameters than its corresponding BERT model, runs 60\% faster but still has 97\% of its performance measured on the GLUE language understanding benchmark \cite{sanh2019distilbert}.

As with ELMo, DistilBERT has been implemented as an embedding layer into our models. However, the texts are already encoded into input ids and attention masks beforehand. We use a maximum text length of 800 and vector length of 256 of task 1 and a maximum text length of 100 and vector length of 32 for task 2. These values have been chosen experimentally and for task 1 gives us the best results before running into memory and training time issues. One should always make sure to use a long enough text length for the input, otherwise important information could be cut and not used for the classification. For example if we have a long newspaper article in task 1 and the protest is only mentioned at the end, the model is not able to make the correct prediction as the information is never available. The input ids and attention masks are then used as the input for the model to the first layer.

\section{Implementation of the models}
\label{section:implementation_of_the_models}
The codebase for model 1 has been kindly provided by Ali Safaya \cite{safaya_event_nodate} and for model 2 by Benjamin J. Radford \cite{radford_multitask_nodate} originally used for the CLEF ProtestNews tasks of 2019.

Both models have been implemented in Python with the help of various libraries. The pandas library\footnote{https://pandas.pydata.org/} has been used for reading and manipulating the datasets provided in the JSON format. Pandas itself relies on the library NumPy\footnote{https://numpy.org/}, which provides many functions for data manipulation based on compiled C code. This makes both NumPy and Pandas very performant compared to plain python code, especially for big datasets.

The preprocessing of the texts relies mostly on the Natural Language Toolkit NLTK\footnote{https://www.nltk.org/}. NLTK provides many useful tools when working with natural language such as tokenization, part-of-speech tagging, lemmatization and stopwords lists.

The resampling of the datasets has been done with the SMOTE implementation of the imbalanced-learn library\footnote{https://imbalanced-learn.org/stable/index.html}.

The neural network models have been implemented with the deep learning framework Keras\footnote{https://keras.io/}. Keras uses the machine learning and artificial intelligence library TensorFlow\footnote{https://www.tensorflow.org/} developed by Google Brain\footnote{https://research.google/teams/brain/} as a backend and provides an interface for building both simple and more complex models. The models are built by constructing either already supported or custom made sequential layers, each taking the output of the last layer as their new input. Keras provides functions for fitting models to input data, specifying the number of epochs and batch sizes per epoch, controlling the performance of the models during the fitting step, making predictions on new data among many others.

Finally, scikit-learn\footnote{https://scikit-learn.org/} has been used for evaluating the models based on various performance metrics such as precision, recall, F1-Score or confusion matrix.

\section{Training the models}
\label{section:training_the_models}
When we built our first model with ELMo, we tried to first directly convert the texts of our training dataset to vectors with ELMo and feed them to the models as input. However, we quickly ran into memory issues running the program locally on a graphics card with 8GB of GDDR5 memory, even when limiting the input size. Instead, we implemented ELMo directly into the model as an embedding layer directly after the input with Keras. This approach was more efficient, but we still faced some memory issues with the GPU.

We found a solution with Google Colaboratory (colab), which allows anyone to write and run python code directly in the browser and already comes with the most commonly used libraries for machine learning tasks installed. The code is not run locally in the browser, but rather in cloud-servers of google. The main advantage for us was that it also provided free use of a randomly assigned graphics card in the cloud. Depending on how many people were currently using the service, we always got assigned either an NVIDIA Tesla K80 GPU with 24GB of GDDR5 memory or the faster NVIDIA T4 GPU with 16GB of GDDR6 memory. The only drawback of google colab was the usage limitation of the GPUs with a time of up to 12 hours per day, but sometimes significantly less if too many people were using the service at the time. 

Besides the GPU, the time it took to fit a model to the training dataset depends on the complexity of the Keras layers, the number of epochs the models were trained and the batch size of the texts that were evaluated together. For our experiments, we trained around 100 models and it took on average around 2 hours to train a model. The predictions of the testing datasets in comparison only took a few minutes.

\chapter{Evaluation and Results}
\label{chapter:evaluation_and_results}
This chapter shows the results and evaluations of the best experiments for each model, task and word embedding. Presenting all the results of every experiment done would go beyond the scope of this chapter, the rest of the results can be found in the Appendix \ref{appendix:A}.

Section \ref{section:evaluation_metrics} Evaluation Metrics shortly explains all the evaluation techniques used in this chapter. Accuracy metrics such as Precision, Recall and macro averaged F1-score metrics were used for evaluating task 1 and task 2 to get an overview of how well the models perform. The next sections show the detailed evaluations of the results for the different experiments. Section \ref{section:analysis_of_the_predictions} Analysis of the Predictions shows some insights to where misclassifications most frequently occured and the reasoning behind it, before summarizing the best results for ELMo and DistilBERT in comparison to the base models in section \ref{section:overview_best_results} Overview of the best Results.

\section{Evaluation Metrics}
\label{section:evaluation_metrics}
For evaluating the models, Ali Hürriyetoğlu kindly setup a competition on CodaLab  (Additional Scoring) CLEF 2019 Lab ProtestNews\footnote{https://competitions.codalab.org/competitions/34242} where we could upload our submissions. For the India\_Test and China\_Test datasets we are limited to the evaluation metrics Precision, Recall and F1-Score because they were evaluated online and we did not have access to the true labels of the test datasets. We utilized the labelled Dev\_Test datasets that were not used in the training process to construct confusion matrices.

\subsection{Confusion Matrix}
The use of the confusion matrix and its metrics in our study shows the issue of class imbalance in the dataset, namely the non-equal number of positive and negative samples. The confusion matrix shows us which classifications were the most difficult and where the misclassifications occurred \cite{crestani_overview_2019}. 

The confusion matrix shows the number of True Positive, True Negative, False Positive and False Negative predictions as follows:

\begin{tabular}{l|l|c|c|c}
\multicolumn{2}{c}{}&\multicolumn{2}{c}{Predicted Class}&\\
\cline{3-4}
\multicolumn{2}{c|}{}&Non-Protest&Protest&\multicolumn{1}{c}{Total}\\
\cline{2-4}
\multirow{2}{*}{Actual Class}& Non-Protest & $TN$ & $FP$   & $TN+FP$\\
\cline{2-4}
&  Protest & $FN$ & $TP$ & $FN+TP$\\
\cline{2-4}
\multicolumn{1}{c}{} & \multicolumn{1}{c}{Total} & \multicolumn{1}{c}{$TN+FN$} & \multicolumn{    1}{c}{$FP+TP$} & \multicolumn{1}{c}{$N=TP+FN+FP+TN$}\\
\end{tabular}\\

\begin{itemize}
  \item \textbf{True Positive (TP)}: The true label is positive (1=Protest) and the predicted label is also positive (1=Protest).
  \item \textbf{True Negative (TN)}: The true label is negative (0=Non-Protest) and the predicted label is also negative (0=Non-Protest).
  \item \textbf{False Positive (FP)}: The true label is negative (0=Non-Protest) but the predicted label is positive (1=Protest).
  \item \textbf{False Negative (FN)}: The true label is positive (1=Protest) but the predicted label is negative (0=Non-Protest).
\end{itemize}

\subsection{Accuracy}
The accuracy (\ref{equation:accuracy}) is the ratio between the correctly predicted samples and the total number of samples. It shows how often the model makes a correct classification.

\begin{equation} 
\label{equation:accuracy}
\begin{split}
Accuracy  & = \frac{TP+TN}{N=TP+FN+FP+TN} 
\end{split}
\end{equation}

\subsection{Macro Average Precision}
Precision is a metric that shows which proportion of the positive predictions were correct and is calculated by dividing the number of True Positive samples by the sum of the True- and False Positive samples (\ref{equation:precision}). Instead of the overall precision we calculate the macro average precision, which is the average precision value between all classes. As opposed to the micro average, the macro average is not weighted, which means the models get punished more for having a low precision value for the minority class, even when this class does not contain many samples.

\begin{equation} 
\label{equation:precision}
\begin{split}
Precision = \frac{TP}{TP+FP}
\end{split}
\end{equation}

\subsection{Macro Average Recall}
Recall calculates what proportion of actual positives were predicted correctly by dividing the number of True Positive samples by the sum of the True Positive and False Negative Samples (\ref{equation:recall}). We again calculate the macro average of both classes.

\begin{equation} 
\label{equation:recall}
\begin{split}
Recall = \frac{TP}{TP+FN}
\end{split}
\end{equation}

\subsection{Macro Average F1-Score Measure}
The F1-Score is the harmonic mean of the Precision and Recall as shown in Equation \ref{equation:f1_score}. Compared to the accuracy, the F1-Score is more suitable for datasets with an imbalanced class distribution as an overall performance metric. As we have seen in Tables \ref{table:task1} and \ref{table:task2}, our training datasets consist of about 80\% non-protest and 20\% protest related articles and sentences. It would be easy to build a model that reaches 80\% accuracy, as it would simply need to always predict the class non-protest. Since the F1-Score is calculated based on the Precision and Recall, it also takes False Positives and False Negatives directly into account. Similar to the Precision and Recall metrics, we also calculate the macro average F1-Score to treat both classes equally. 

\begin{equation} 
\label{equation:f1_score}
\begin{split}
F1-Score & = \frac{2*Precision*Recall}{Precision+Recall}
\end{split}
\end{equation}

\section{Task 1 Results for Model 1.1}

For the binary classification task of the news articles, the first ELMo and DistilBERT embedding models were implemented with 1-Dimensional Convolution layers and a MaxPooling1D layers (see \ref{subsection:model1:architecture} ELMo and DistilBERT Architecture for Model 1).

\begin{figure}[H]
\begin{minipage}{.52\textwidth} 
    \includegraphics[width=1\textwidth]{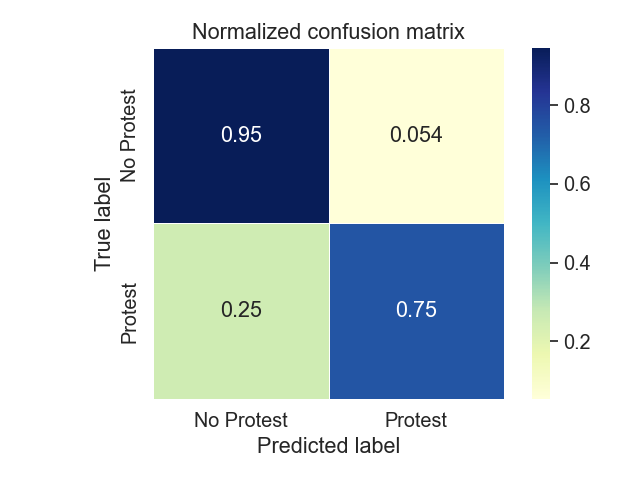}
    \caption{Task 1 - Model 1.1 ELMo.}
    \label{ELMo_Task1_SAF}
\end{minipage}%
\begin{minipage}{.52\textwidth}
    \includegraphics[width=1\textwidth]{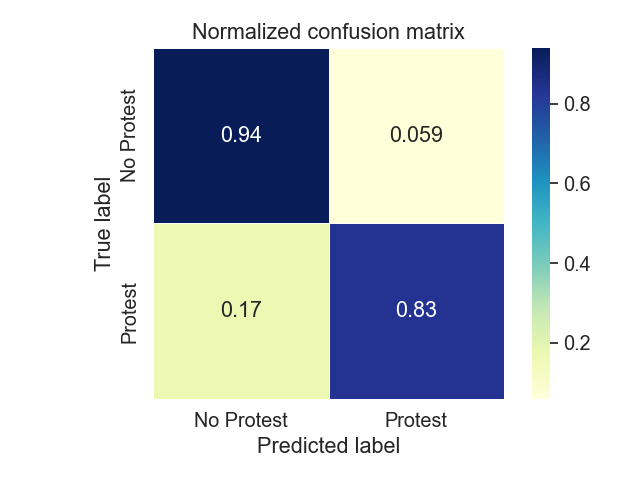}\hfill
     \caption{Task 1 - Model 1.1 DistilBERT.}
     \label{DistilBERT_Task1_SAF}
\end{minipage}
\end{figure}

The best ELMo model for this task got an accuracy of 90\% for the Dev\_Test dataset. The confusion matrix in Figure \ref{ELMo_Task1_SAF} shows that the model predicts the non-protest class very well with a normalized True Negative value of 0.95 and a False Positive value of 0.05. It is immediately visible that the prediction for the minority class protest is more difficult with a True Positive value of 0.75. It has a bias towards the prediction of the majority class non-protest, falsely predicting non-protest news in 25\% of cases.

The best DistilBERT model in comparison has a similar True Negative value of 0.94, also finding the non-protest news articles reliably. Furthermore, it also finds the protest news articles more reliably with a True Positive rate of 0.83. It is less biased towards predicting the majority class shown by the smaller False Negative rate of 0.17.

The Table \ref{table:task1_Model1.1_ELMoDBERTExpSAF} shows an overview of the different data cleaning procedures for both the ELMo and DistilBERT model for all the different datasets. Both models perform fairly well for the Dev\_Test and India\_Test datasets, the news articles of which come from the same country India as the training dataset. We can see a clear decrease in performance for the China\_Test dataset based on all metrics. For the dataset containing news articles from China we can also see the advantage of the datacleaning. In both models lightly cleaning and cleaning the dataset leads to a better performance for the China\_Test dataset compared to the not cleaned dataset. For ELMo, the lightly cleaned dataset reached the highest average F1-Score of 0.65, while for DistilBERT the cleaned dataset got the highest average F1-Score of 0.71.

\begin{table}[H]
\begin{center}
\centering
\caption{Model 1.1: ELMo and DistilBERT - Text Classification.}
\label{table:task1_Model1.1_ELMoDBERTExpSAF}
\begin{tabular}{|c|c|c|c|c|c|c|c|c|c|c|c|c|}
\hline 
\textbf{\multirow{1}{*}{Task 1}} &
\multicolumn{3}{c|}{\textbf{Dev\_Test}} & \multicolumn{3}{c|}{\textbf{India\_Test}} & \multicolumn{3}{c|}{\textbf{China\_Test}} & \multicolumn{3}{c|}{\textbf{Avg. Results}} \\
\hline 
\cline{1-13} 
\textbf{{ELMo}} & \textbf{{P}} & \textbf{{R}} & \textbf{{F1}} & \textbf{{P}} & \textbf{{R}} & \textbf{{F1}} & \textbf{{P}} & \textbf{{R}} & \textbf{{F1}} & \textbf{{P}} & \textbf{{R}} & \textbf{{F1}} \\\hline
\textbf{LightClean}  & 0.86   &  0.85  & \textbf{0.85} &	0.75 &	0.75 &	\textbf{0.75} &	0.69 &	0.5 &\textbf{0.55} & 0.69 &	0.62 &	\textbf{0.65}  \\
\hline
\textbf{Clean}  & 0.86 &  0.84 & 0.85 &  0.73 & 0.73 & 0.73 &	0.35 &	0.57 &	0.43 &	0.54 &	0.65 &	0.58   \\
\hline
\textbf{Not\_Clean}  & 0.92 & 0.84 &	0.87 & 0.84 &	0.70 &	0.76 &	0.54 &	0.26 &
0.35 & 0.69 &	0.48 &	0.55 \\
\hline
\cline{1-13} 
\textbf{{DistilBERT}} & \textbf{{P}} & \textbf{{R}} & \textbf{{F1}} & \textbf{{P}} & \textbf{{R}} & \textbf{{F1}} & \textbf{{P}} & \textbf{{R}} & \textbf{{F1}} & \textbf{{P}} & \textbf{{R}} & \textbf{{F1}} \\\hline
\cline{1-13}
\textbf{LightClean} & 	0.89 & 0.89 &	0.89 &	0.79 &	0.86 &	0.82 &	0.59 &	0.51 & 	0.55 &	0.69 &	0.68 & 	0.68   \\
\hline
\textbf{Clean} & 0.88 & 0.89  & \textbf{{0.88}} &	0.78 &	0.86 &	\textbf{{0.82}} &	0.57 &	0.64 &	\textbf{{0.60}} &	0.68 &	0.75 &	\textbf{{0.71}} \\
\hline
\textbf{NotClean} & 0.90   &  0.86 &  0.88 &	0.89 &	0.73 &	0.80 &	0.61 &	0.45 &	0.52 &	0.75 &	0.59 & 0.66   \\
\hline
\end{tabular}
\end{center}
\end{table}

\section{Task 2 Results for Model 1.2}

\begin{figure}[H]
\begin{minipage}{.52\textwidth} 
    \includegraphics[width=1\textwidth]{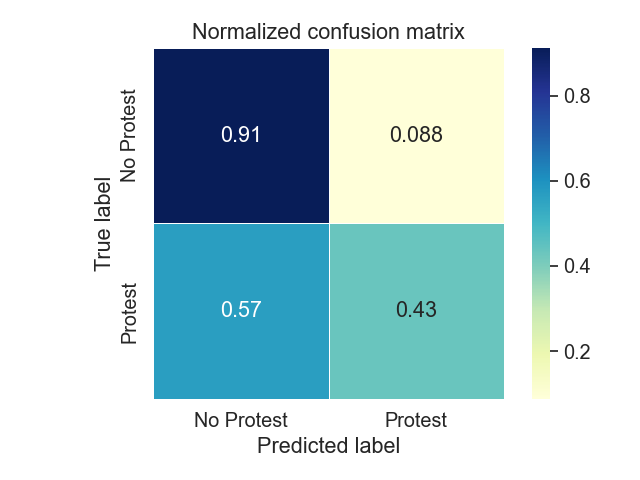}
    \caption{Task 2 - Model 1.2 ELMo.}
    \label{ELMo_Task2_SAF}
\end{minipage}%
\begin{minipage}{.52\textwidth}
    \includegraphics[width=1\textwidth]{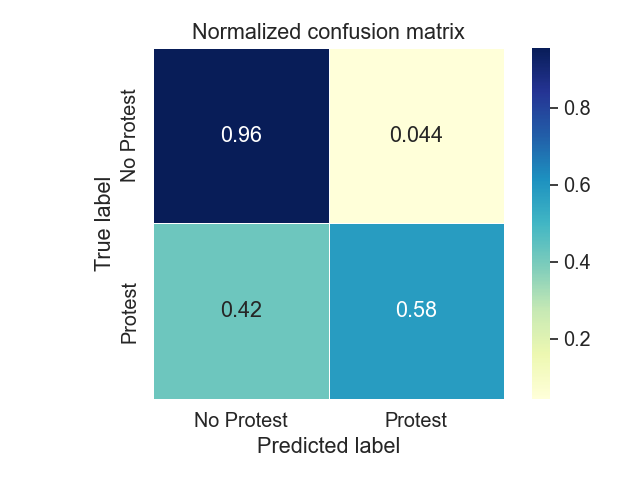}\hfill
     \caption{Task 2 - Model 1.2 DistilBERT.}
     \label{DistilBERT_Task2_SAF}
\end{minipage}
\end{figure}

The first models of ELMo and DistilBERT for Task 2 were implemented with a combination of Bidirectional Gated Recurrent Units and Attention Layers (see Figure \ref{ELMO-DistilBERT_Architecture_SAF}). The Figures \ref{ELMo_Task2_SAF} and \ref{DistilBERT_Task2_SAF} again show the confusion matrices of the best model for the development test datasets. We can see that both the ELMo and DistilBERT models reliably find the non-protest news sentences with a True Negative value of 0.91 and 0.96 respectively. Even though the ELMo model reaches an accuracy of 81\% on the Dev\_Test dataset, it predicts more False Negatives (0.57) than True Positives (0.43), which means it is unable to find the protest-related sentences in most cases. The DistilBERT model performs better for both non-protest and protest related sentences, reaching an accuracy of 88\% with a True Negative rate of 0.96 and a True Positive rate of 0.58. 

The Table \ref{table:task2_Model1.2_ELMoDBERTExpSAF} confirms the overall better performance of the DistilBERT model, especially for the sentences from Indian news articles. The F1-Score of the cleaned datasets are very stable over the different contexts (India and China), indicating that cleaning the dataset makes the model more generalizable. For ELMo, the cleaned dataset had also the best overall F1-Score due to the weak performance of the lightly cleaned and not cleaned datasets for China\_Test.

Interestingly, the uncleaned dataset got the highest F1-Score for both the India and China datasets for the DistilBERT model. This indicates that DistilBERT word embeddings could extract the meaning of a sentence without relying on context-specific information, making  a further cleaning of the sentences counterproductive as some information is lost during the process.

\begin{table}[H]

\begin{center}
\centering
\caption{Model 1.2: ELMo and DistilBERT - Event Sentence Detection.}
\label{table:task2_Model1.2_ELMoDBERTExpSAF}
\begin{tabular}{|c|c|c|c|c|c|c|c|c|c|c|c|c|}
\hline 
\textbf{\multirow{1}{*}{Task 2 }} &
\multicolumn{3}{c|}{\textbf{Dev\_Test}} & \multicolumn{3}{c|}{\textbf{India\_Test}} & \multicolumn{3}{c|}{\textbf{China\_Test}} & \multicolumn{3}{c|}{\textbf{Avg. Results}} \\
\hline 
\cline{1-13} 
\textbf{{ELMo}} & \textbf{{P}} & \textbf{{R}} & \textbf{{F1}} & \textbf{{P}} & \textbf{{R}} & \textbf{{F1}} & \textbf{{P}} & \textbf{{R}} & \textbf{{F1}} & \textbf{{P}} & \textbf{{R}} & \textbf{{F1}} \\\hline
\textbf{LightClean}  &  0.69   &   	0.66  &  0.67  &     	0.43 &	0.39 &	0.41 &	0.38 & 	0.25 &	0.30 &	0.40 &	0.32 &	0.35	  \\
\hline
\textbf{Clean}  & 0.71  &   0.67 &   \textbf{0.69}  &  0.42 &	0.48 &	\textbf{0.44} &	0.45 &	0.42 &	\textbf{0.44} &	0.44 &	0.45 &	\textbf{0.44}  \\
\hline
\textbf{NotClean}  & 0.73    &  	0.70 &	0.71  &     	0.45 &
0.42 &	0.44 &	0.49 &	0.28 &	0.36 &	0.47 &	0.35 &	0.39 \\
\hline
\cline{1-13} 
\textbf{{DistilBERT}} & \textbf{{P}} & \textbf{{R}} & \textbf{{F1}} & \textbf{{P}} & \textbf{{R}} & \textbf{{F1}} & \textbf{{P}} & \textbf{{R}} & \textbf{{F1}} & \textbf{{P}} & \textbf{{R}} & \textbf{{F1}} \\
\hline
\textbf{LightClean} & 0.84   &   	0.78  &    	0.80 &	0.71 &	0.57 &	0.63 &	0.62 &	0.26 &	0.37 &	0.66 &	0.41 &	0.50 \\
\hline
\textbf{Clean} & 0.67 & 0.75 & 0.67 &	0.28 &	0.76 &	0.40 &	0.25 &	0.76 &	0.37 &	0.26  &	0.76 &	0.39\\
\hline
\textbf{NotClean} & 0.84 & 	0.77 & 	 \textbf{0.79} & 	0.74 &	0.61 &	 \textbf{0.66} &	0.62 &	0.36 &	0.46 & 0.68 & 0.49 &  \textbf{0.56} \\
\hline
\end{tabular}
\end{center}
\end{table}

\section{Task 1 and Task 2 Results for Model 2}

The second model is based on Bidirectional Long Short-Term Memory for both tasks, the architecture has been described in detail in \ref{subsection:architecture_model2} ELMo and DistilBERT Architecture for Model 2. 

Unfortunately, we were unable to get a performance worth evaluating when implementing this model with ELMo word embeddings. The model always predicted the majority class, even when trying to resample the datasets. The only way of getting predictions for both classes was to change the architecture of the model. However, this was not the goal of this thesis which is why we only evaluate Model 2 with DistilBERT word embeddings.

\begin{figure}[H]
\begin{minipage}{.52\textwidth} 
    \includegraphics[width=1\textwidth]{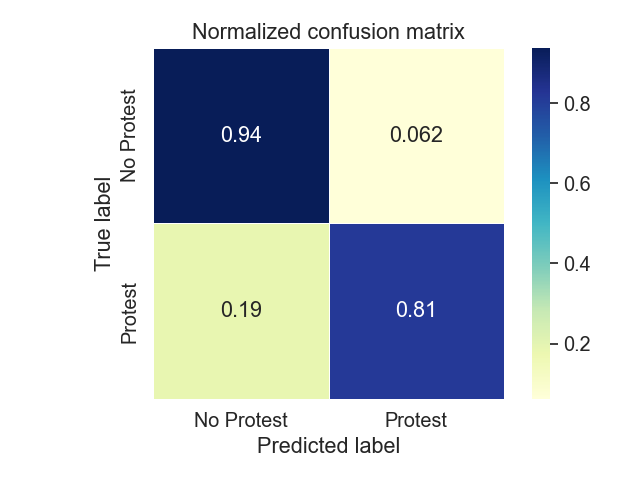}
    \caption{Task 1 - Model 2 DistilBERT.}
    \label{DistilBERT_Task1_BEN}
\end{minipage}%
\begin{minipage}{.52\textwidth}
    \includegraphics[width=1\textwidth]{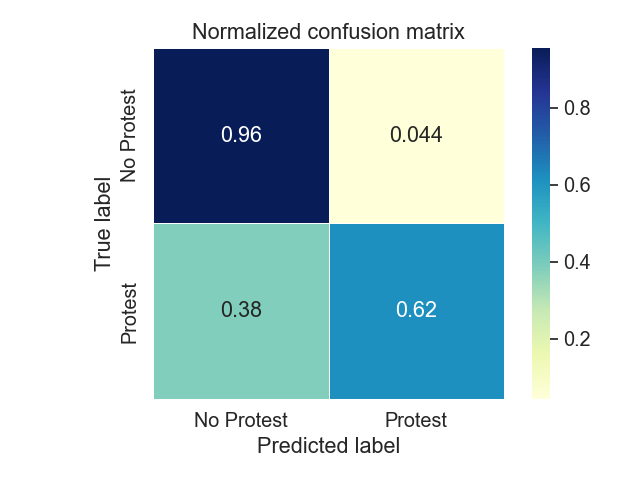}\hfill
     \caption{Task 2 - Model 2 DistilBERT.}
     \label{DistilBERT_Task2_BEN}
\end{minipage}
\end{figure}

Figure \ref{DistilBERT_Task1_BEN} shows the confusion matrix of the best model for task 1 measured on the Dev\_Test dataset. As with model 1, the model reliably finds non-protest news articles with a True Negative value of 0.94. As can be expected, finding the minority class of protest related news articles is more difficult, with a True Positive rate of 0.81. The accuracy reaches 91\% due to the majority of the samples being in the non-protest class.

Figure \ref{DistilBERT_Task2_BEN} for task 2 of the Dev\_Test dataset draws a similar picture. Even though the accuracy of the model is 89\% with a good True Negative value of 0.96, we can see the difficulty in finding the protest related sentences on the True Positive value of 0.62. 

Table \ref{table:task12_Model2_DBERTExpBEN} shows the results on both tasks with the three different data cleaning methods. The cleaned dataset helped the models perform the best across context visible in the F1-Score of the China\_Test dataset for both tasks. The recall measure is often times the deciding factor, resulting in the best F1-Score even when the precision is comparably lower. For task 1 we got the best average F1-Score with the lightly cleaned dataset, performing well on both the India\_Test and China\_Test datasets. For task 2 the cleaned dataset got the highest average F1-Score over both test datasets, outperforming the lightly cleaned and not cleaned dataset in both the India and China dataset.

\begin{table}[H]
\begin{center}
\centering
\caption{Model 2: DistilBERT - Text Classification and Event Sentence Detection.}
\label{table:task12_Model2_DBERTExpBEN}
\begin{tabular}{|c|c|c|c|c|c|c|c|c|c|c|c|c|}
\hline 
\textbf{\multirow{1}{*}{DistilBERT}} &
\multicolumn{3}{c|}{\textbf{Dev\_Test}} & \multicolumn{3}{c|}{\textbf{India\_Test}} & \multicolumn{3}{c|}{\textbf{China\_Test}} & \multicolumn{3}{c|}{\textbf{Avg. Results}} \\
\hline
\cline{1-13} 
\textbf{{Task 1}} & \textbf{{P}} & \textbf{{R}} & \textbf{{F1}} & \textbf{{P}} & \textbf{{R}} & \textbf{{F1}} & \textbf{{P}} & \textbf{{R}} & \textbf{{F1}} & \textbf{{P}} & \textbf{{R}} & \textbf{{F1}} \\
\hline
\textbf{LightClean}  & 0.87     & 	0.88 &     	\textbf{0.87} &	0.81 &	0.84 &	\textbf{0.82} &	0.69 &	0.57 &	\textbf{0.62} &	0.75 &	0.70	& \textbf{0.72} \\
\hline
\textbf{Clean}  & 0.89   &   	0.87  &    	0.88  &     	0.78 &	0.75 &	0.77 &	0.73 &	0.55 &	0.63 &	0.76 &	0.65 &	0.70 \\
\hline
\textbf{NotClean}  & 0.87   &   	0.88 &	0.88 &	0.85 &	0.80 &	0.82 &	0.63 &	0.58 &	0.60 &	0.74 &	0.69 &	0.71\\
\hline
\cline{1-13} 
\textbf{{Task 2}} & \textbf{{P}} & \textbf{{R}} & \textbf{{F1}} & \textbf{{P}} & \textbf{{R}} & \textbf{{F1}} & \textbf{{P}} & \textbf{{R}} & \textbf{{F1}} & \textbf{{P}} & \textbf{{R}} & \textbf{{F1}} \\
\hline 
\textbf{LightClean}  & 0.85   &   	0.76  &    	0.79 &	0.76 &	0.54 &	0.63 &	0.66 &	0.33 &	0.44 &	0.71 &	0.44 &	0.54 \\
\hline
\textbf{Clean}  & 0.85      & 	0.79  &      \textbf{0.81}  &	0.69  &	0.61  &	\textbf{0.65}  &	0.66  &	0.45  &	\textbf{0.54}  &	0.68 &	0.53  &	\textbf{0.59} \\
\hline
\textbf{NotClean}  & 0.85 &      	0.76 &      	0.79 &	0.76 &	0.62 &	0.68 &	0.61 &	0.36 &	0.45 &	0.69 &	0.49 &	0.57 \\
\hline

\end{tabular}
*\textbf{P}  = Precision, \textbf{R} = Recall, \textbf{F1} = F1\_Score
\end{center}
\end{table}

\section{Analysis of the predictions}
\label{section:analysis_of_the_predictions}
Using the labelled development test dataset, we took a closer look at the texts that were commonly misclassified by the models. We were surprised to find words like \textit{protest} and \textit{protesting} even in texts classified as non-protest, resulting in a False Negative. Even though this occurred very rarely and the classfication still depends on the context, this could be a good opportunity to introduce further improvements to the model. One such improvement could be using an "experts opinion" and augment the model, resulting in a hybrid model \cite{gennatas2020expert}. In our case this could for example be done by manually adjusting the feature weight of certain words or partial sentences, which empirically denote either protest or non-protest.

Another reason for misclassification was in task 1, when protest related information was only clearly contained in a single sentence of an otherwise larger news article. Uncommon words relating to protest also proved difficult for the models, for example the word \textit{agitation} was not clearly related to protest news by the model in one case, also resulting in a False Negative prediction.

False Positive predictions happened frequently due to events such as burglaries, which can result in similar descriptions as protest events. Another common pitfall for False Positive predictions was the mentioning of military and or political events, which can often times lead to the article or sentence being about protests. However, even when these texts did not contain information about protests, they were frequently misclassified as such. 

Comparing the uncleaned dataset predictions to the cleaned datasets, we found that the uncleaned dataset model wrongly gave meaning to words that are not relevant to protests. This could be due to coincidences in the way Indian newspapers write about protests, which is not true for other countries (like China in our case).

\section{Overview of the best Results}
\label{section:overview_best_results}
Table \ref{table:task1_2ExperimentComparisons} shows the performance of the best models with ELMo and DistilBERT word embeddings compared to the base models. 

For our implementations of model 1 we notice that DistilBERT performed better in both task 1 and task 2. Compared to the original model 1 that uses word2vec word embeddings, we get a similar performance with DistilBERT for task 1. On the other hand, the DistilBERT model for task 2 performed worse than the base model using word2vec with an average F1-Score of 0.56 compared to 0.65. The overall F1-Score averaged over both tasks for model 1 was accordingly higher with the original word embeddings compared to DistilBERT. 

For model 2 we did get a notable improvement with DistilBERT over FastText word embeddings in task 2 for the China dataset, improving the F1-Score from 0.45 to 0.54. This results in a higher average F1-Score over both tasks of 0.66 compared to 0.65.

\begin{table}[H]
\begin{center}
\centering
\caption{Comparison between ELMo and DistilBERT models results with the best F1-Score results for the CLEF 2019 Lab ProtestNews Models for Binary Classification and Event Sentences Detection Tasks.}
\label{table:task1_2ExperimentComparisons}
\begin{tabular}{|c|c|c|c|c|c|c|c|}
\hline 
\textbf{\multirow{1}{*}{}} &
\multicolumn{3}{c|}{\textbf{Task 1}} &
\multicolumn{3}{c|}{\textbf{Task 2}} &
\multicolumn{1}{c|}{\textbf{Overview}}\\
\hline 
\textbf{\multirow{1}{*}{Model}} &
\multicolumn{1}{c|}{\textbf{Test 1-1}} & \multicolumn{1}{c|}{\textbf{Test 1-2}} &
\multicolumn{1}{c|}{\textbf{Avg.}} &
\multicolumn{1}{c|}{\textbf{Test 2-1}} &
\multicolumn{1}{c|}{\textbf{Test 2-2}} & \multicolumn{1}{c|}{\textbf{Avg.}} &
\multicolumn{1}{c|}{\textbf{F1 Avg. T1-T2}}\\
\hline 
\textbf{Model 1 - Word2vec}  & 0.81   &  0.63  & 0.72 &	0.70 &	0.60 &	\textbf{0.65} &	0.$69^1$   \\
\hline
\textbf{Model 1 - ELMo}  & 0.75   &  0.55  & 0.65 &	0.44 &	0.44 &	0.44 &	0.55  \\
\hline
\textbf{Model 1 - DistilBERT}  & 0.82   &  0.60 & 0.71 &	0.66 &	0.46 & 0.56 &	0.64   \\
\hline
\textbf{Model 2 - FastText}  & 0.83   &  0.65  & 0.74 &	0.66 &	0.45 &	0.56 &	0.$65^1$\\
\hline
\textbf{Model 2 - DistilBERT}  & 0.82  &  0.62 & 0.72 &	0.65 &	\textbf{0.54} &	\textbf{0.59} &	\textbf{0.66} \\
\hline
\end{tabular}
$^1$We got the F1-Score Avg. Results for the test datasets from \cite{crestani_overview_2019}.

*\textbf{Test 1-1, 2-1}  = India Test Results, \textbf{Test 1-2, 2-2}  = China Test Results\\

\end{center}
\end{table}

\chapter{Discussion}
To start of this chapter, we are first going to answer the three research questions proposed in chapter \ref{chapter:introduction} Introduction.

\subsubsection*{Does the use of ELMo word embeddings increase the performance of an already established model using other word embeddings?}
According to the experiments done in the scope of this thesis we answer this question with a no. 

For the first models, we replaced the word2vec word embeddings with ELMo word embeddings without changing anything else about the models. We have experimented with different numbers of epochs, adapted the learning rate and tried out different levels of cleaning the datasets. Our best models with ELMo got an average F1-Score over both tasks of 0.55 compared to 0.69 of the original models using word2vec.

For the second models we were unable to get useful predictions with ELMo instead of FastText, as both models for task 1 and task 2 always predicted the minority class. On this basis we could not do any further evaluation of these models.

\subsubsection*{Does the use of DistilBERT word embeddings increase the performance of an already established model using other word embeddings?}

For the first models originally implemented with word2vec we did not get an improvement when replacing the word embeddings with DistilBERT. Our best implementation of the models with DistilBERT got an average F1-Score over both tasks of 0.55 compared to 0.69 of the base models.

For the second models we replaced the FastText word embeddings with DistilBERT and got an average F1-Score over both tasks of 0.66 compared to 0.65 of the base models. 

To answer the research question, we can not generally say that DistilBERT can be used to replace any existing word embedding method and see an improvement. We did however get an improvement when replacing FastText with DistilBERT without changing anything else about the models.

\subsubsection*{Does ELMo or DistilBERT perform better measured on two well performing models in the CLEF 2019 Protest Event?}

In our experiments we got the best results for both models and tasks with DistilBERT rather than ELMo. Additionally, DistilBERT could be successfully implemented and even improved the second models, while ELMo always predicted the majority class using the same models.\\

\noindent It is important to note at this point that these results do not say anything about which word embedding method is best. We evaluated both ELMo and DistilBERT word embeddings in a very restricted environment by not changing the architecture of the base models other than the word embedding layer. It is clear that different word embedding methods work well with different models, and neither ELMo nor DistilBERT worked best in all cases. However, we have also shown that it is possible to improve a model by changing the word embedding layer and leaving the rest of the architecture as is.

Evaluating the models with a macro-average F1-Score on imbalanced datasets also showed that correctly predicting the minority class was key in getting a good result. There was a trade-off between precision and recall with getting good results in the latter was much more difficult to achieve. All of the best models had a bias towards predicting the majority class, which makes sense as predicting the class that appears more often will be the correct choice most of the time. A higher bias can easily be detected based on a lower recall measure, which in turn decreases the F1-Score. From a practical standpoint it makes sense in this case to choose such evaluation metrics that punish a bad performance in the minority class more than to reward a good performance in the majority class. It is more important to find the few protest news related articles among all the other news articles than it is to, for example, get a high accuracy score.

One of the goals of the CLEF 2019 tasks was to build a model that generalizes well outside of the country domain of the training dataset, hence the testing dataset of China. We have found that both ELMo and DistilBERT generalize better when preprocessing the dataset beforehand. The only exception was the DistilBERT implementation for the protest sentence detection in task 2 with the second model, where the uncleaned datasets performed best even for the China dataset. Between ELMo and DistilBERT the data is inconclusive as to which word embedding method generalizes better across the country domain.

\chapter{Conclusion and Future Work}
The goal of this thesis was to find out if the word embedding methods ELMo and DistilBERT could be implemented into existing models, replacing other word embeddings and getting a better performance without changing anything else about the model's architecture. Using two well performing models from the CLEF 2019 ProtestNews tasks 1 and 2, we replaced the existing word embeddings word2vec and FastText with ELMo and DistilBERT respectively. The word embeddings were integrated into the models as a Keras layer and trained with the use of a GPU on google colab. 

The datasets have been resampled with SMOTE to counteract the imbalance in the two classes protest and non-protest news related articles and sentences. This was crucial for the DistilBERT word embeddings, as the loss of the models during training were otherwise not converging towards 0. The implementation of ELMo into the second model without changing the architecture was not successful as it always predicted the majority class non-protest.

The datasets were preprocessed three different ways, once without any data cleaning, once with a light cleaning such as removing stopwords and once with a more thorough data cleaning while also lemmatizing the words. Cleaning the datasets helped the models generalize across the country, as the training datasets were from India and the testing datasets also included news article datasets from China. Only one model on the second task of protest sentence detection performed better on an uncleaned dataset with DistilBERT.

The replacement of word2vec with ELMo and DistilBERT in the first model did not lead to a better average F1-Score across the two tasks, even with DistilBERT clearly outperforming ELMo. For the second model, DistilBERT got an overall better performance than the baseline model using FastText. Especially for the China dataset of Task 2, DistilBERT reached an F1-Score of 0.54 compared to the baseline of 0.45.

Neither ELMo nor DistilBERT proved to be the best choice across all models compared with the original word embeddings, nevertheless between the two, DistilBERT showed more promise by improving the performance of the second model. It would be interesting to see if DistilBERT could improve other models, possibly by lowering the restrictions and allowing small changes to the architecture. Keeping the models exactly the same apart from the word embeddings may be too restrictive. It has already been shown that DistilBERT (and ELMo) can perform very well on these same tasks with the right model \cite{buyukoz_analyzing_nodate}.\\

\noindent Further work could be done through adjusting the feature weight of the word embeddings by training the word embedding models on a sufficiently large dataset in the same context of this task instead of using pretrained models. Another possibility would be an adjustment by an expert, giving more weight to terms and meanings clearly relating to protest or non-protest, resulting in a hybrid model. 

\bibliography{thesis}
\bibliographystyle{plain}

\appendix

\begin{appendices}
\chapter{Appendix}
\label{appendix:A}
\section{Aditional information about other Experiment results running with ELMo for Model 1.}

\begin{figure}[H]
    \centering
    \includegraphics[width=1.1\textwidth]{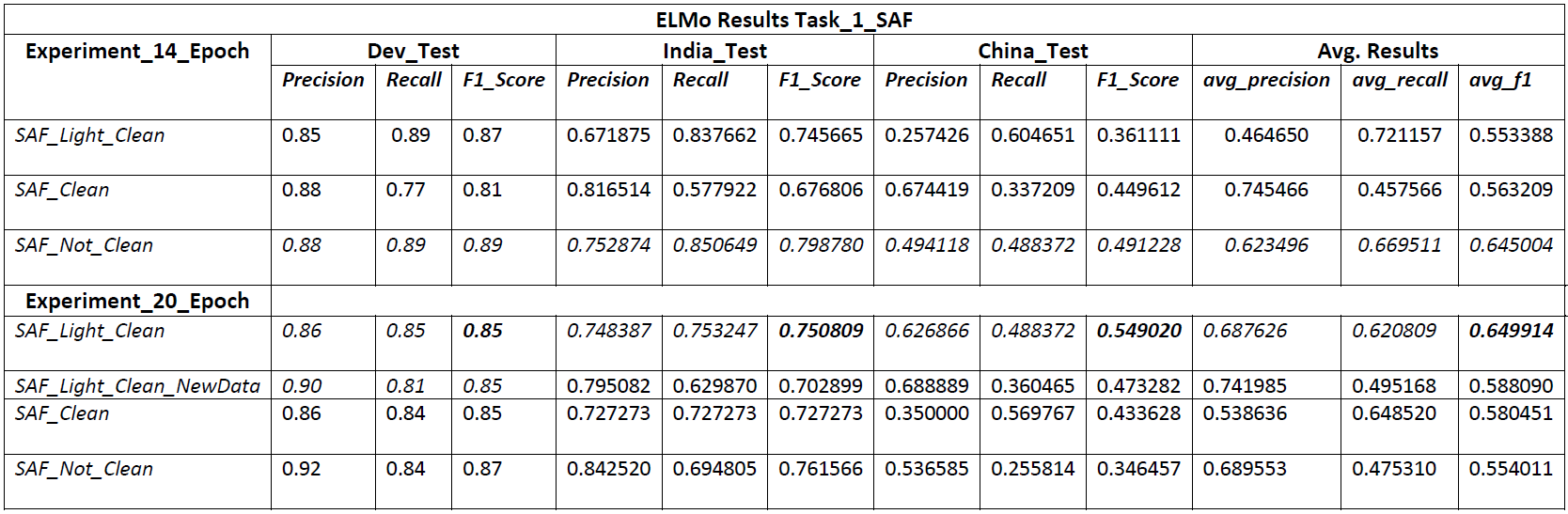}\hfill
     \caption{Task 1 - ELMo results for Model1.}
     \label{ELMO-DistilBERT_Architecture_BEN1}
\end{figure}

\begin{figure}[H]
    \centering
    \includegraphics[width=1.1\textwidth]{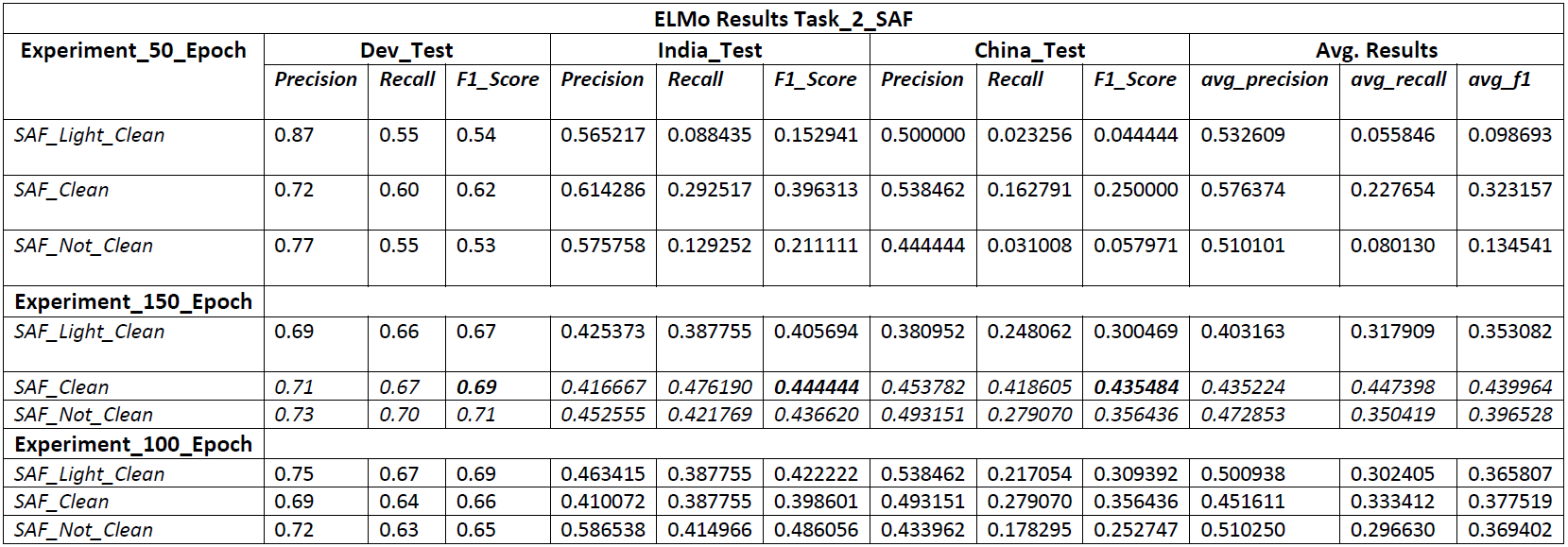}\hfill
     \caption{Task 2 - ELMo results for Model1.}
     \label{ELMO-DistilBERT_Architecture_BEN2}
\end{figure}

\label{appendix:B}
\section{DistilBERT Experiment results for Model 1.}

\begin{figure}[H]
    \centering
    \includegraphics[width=1.1\textwidth]{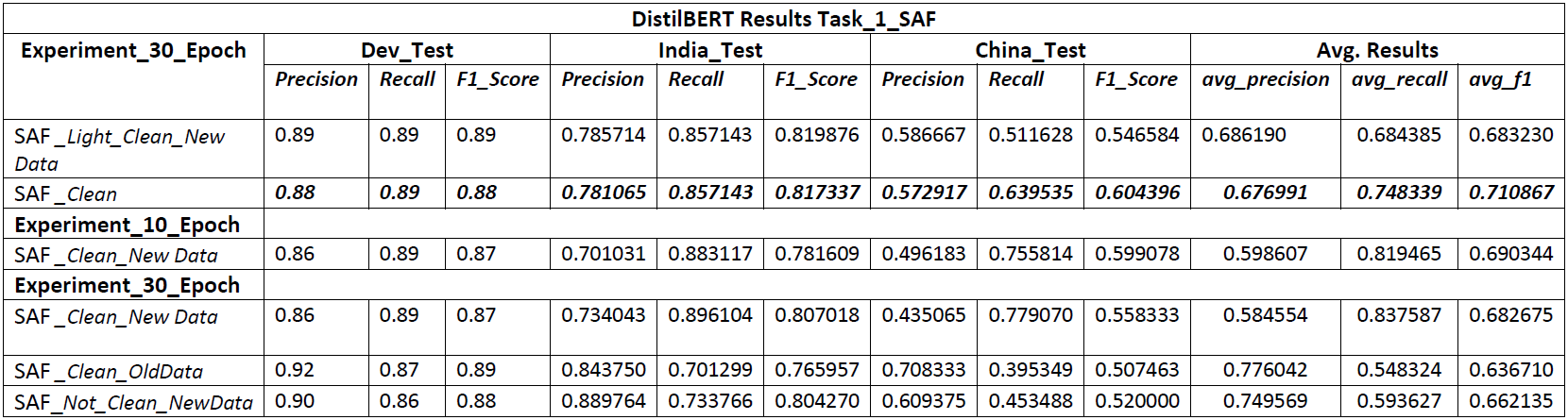}\hfill
     \caption{Task 1 - DistilBERT results for Model 1.}
     \label{ELMO-DistilBERT_Architecture_BEN3}
\end{figure}

\begin{figure}[H]
    \centering
    \includegraphics[width=1.1\textwidth]{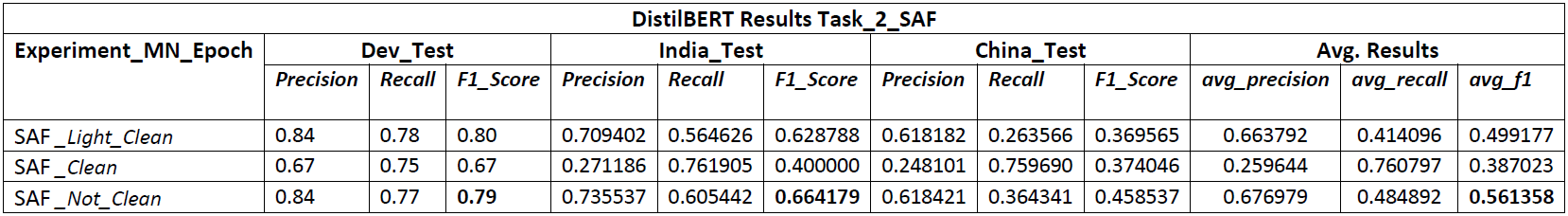}\hfill
     \caption{Task 2 - DistilBERT results for Model 1.}
     \label{ELMO-DistilBERT_Architecture_BEN4}
\end{figure}

\section{DistilBERT Experiment results for Model 2.}

\begin{figure}[H]
    \centering
    \includegraphics[width=1.1\textwidth]{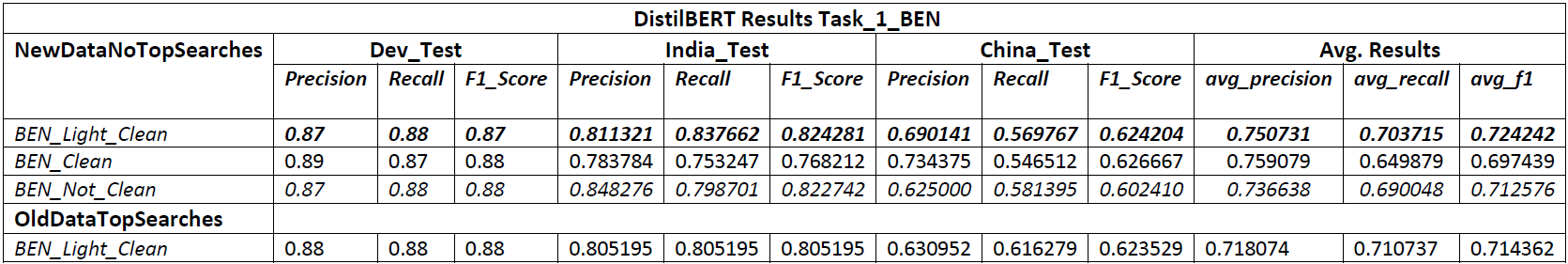}\hfill
     \caption{Task 1 - DistilBERT results for Model 2.}
     \label{ELMO-DistilBERT_Architecture_BEN5}
\end{figure}

\begin{figure}[H]
    \centering
    \includegraphics[width=1.1\textwidth]{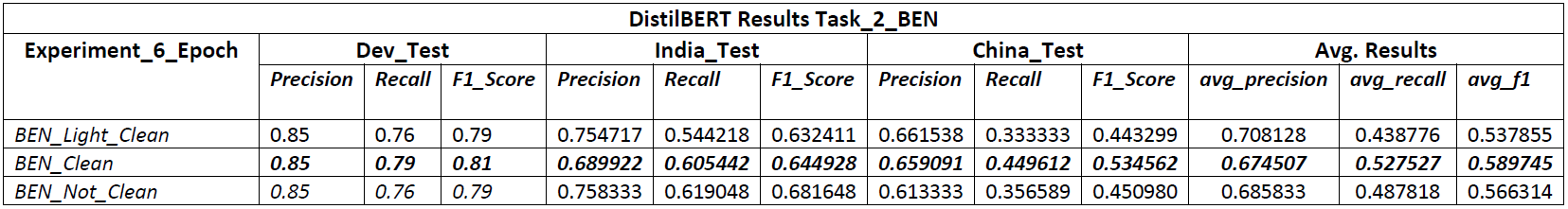}\hfill
     \caption{Task 2 - DistilBERT results for Model 2.}
     \label{ELMO-DistilBERT_Architecture_BEN6}
\end{figure}

\end{appendices}

\end{document}